%% file: main.tex
\documentclass{nature} 

\usepackage[T1]{fontenc}
\usepackage[utf8]{inputenc}
\usepackage{textgreek}

\usepackage{helvet}    
\usepackage{sectsty}   
\allsectionsfont{\sffamily}  

\usepackage{newtxtext,newtxmath}    
\usepackage{amsmath, amssymb, bm} 
\usepackage{amsthm}               
\usepackage{graphicx}             
\usepackage{xspace}               
\usepackage{booktabs}             
\usepackage{multirow}
\usepackage{pdflscape}            
\usepackage[left=1in,right=1in,bottom=1.1in,top=1in]{geometry}
\usepackage{enumitem}             
\usepackage[table,dvipsnames]{xcolor} 
\usepackage{siunitx}              
\usepackage{textcomp}
\usepackage[normalem]{ulem}       
\usepackage[ruled,vlined,linesnumbered]{algorithm2e} 
\usepackage[innercaption]{sidecap} 
\usepackage{lineno}               
\usepackage{verbatim}             
\let\spacing\relax
\usepackage{setspace}             
\usepackage{xr}                   
\usepackage[colorlinks,citecolor=blue,urlcolor=blue]{hyperref} 
\usepackage{xr-hyper}
\usepackage{float}
\usepackage{url}
\usepackage{orcidlink}
\usepackage{threeparttable}

\usepackage{array}
\usepackage{tabularx}
\usepackage{xltabular}

\newcolumntype{L}[1]{>{\raggedright\arraybackslash}p{#1}}
\newcolumntype{C}[1]{>{\centering\arraybackslash}p{#1}}
\newcolumntype{Y}{>{\raggedright\arraybackslash}X}

\usepackage{longtable}
\usepackage{threeparttablex}

\DeclareSIUnit\Molar{M}
\DeclareSIUnit\rpm{rpm}
\DeclareSIUnit\ppm{ppm}

\usepackage[natbib=true, citestyle=numeric-comp,
            bibstyle=nature, sorting=none,
            maxbibnames=6, doi=true, url=false, isbn=false, date=year]{biblatex}

\AtEveryBibitem{%
  \clearfield{note}%
  \clearlist{note}%
  \clearfield{language}%
  \clearlist{language}%
}

\usepackage[font=footnotesize, labelfont={sf,bf}]{caption}
\DeclareCaptionLabelFormat{extended_data}{\textbf{Extended Data #1 #2: }}

\graphicspath{{FIG/}}

\newcommand{\revision}[1]{#1}

\newcommand{\xhdr}[1]{\vspace{1.2mm}\noindent\textsf{\textbf{{#1.}}}\xspace}

\newcommand{\eg}{\emph{e.g.}\xspace}
\newcommand{\ie}{\emph{i.e.}\xspace}

\newcommand{\textsbf}[1]{\textsf{\textbf{#1}}}

\SetKwRepeat{Do}{do}{while}
\let\oldnl\nl
\newcommand{\nonl}{\renewcommand{\nl}{\let\nl\oldnl}}

\theoremstyle{definition}

\newcommand\kg{\textsc{OptimusKG}\xspace}
\newcommand\numnodes{190,939\xspace}
\newcommand\numedges{21,818,752\xspace}
\newcommand\numnodetypes{10\xspace}
\newcommand\numedgetypes{27\xspace}
\newcommand\numpropertyinstances{67,070,490\xspace}
\newcommand\numpropertyvalues{109,665,797\xspace}
\newcommand\numpropertykeys{145\xspace}

\newcommand\numontologies{18\xspace}

\usepackage{soul}
\usepackage[most]{tcolorbox}
\usepackage{xparse}
\usepackage{seqsplit}

\definecolor{codebg}{gray}{0.95}
\sethlcolor{codebg}
\NewDocumentCommand{\code}{m o}{%
  {\fontsize{11}{10}\selectfont\texttt{\seqsplit{#1}\IfValueT{#2}{:\seqsplit{#2}}}}%
}

\newtcolorbox{codeblock}{
  enhanced,
  breakable,
  colback=gray!5,
  colframe=gray!40,
  boxrule=0.4pt,
  arc=3pt,
  left=4pt, right=4pt, top=4pt, bottom=4pt,
  boxsep=0pt,
  fontupper=\ttfamily,
}

\title{
\begin{center}
Unifying biomedical knowledge \\in a modern multimodal graph
\vspace{-10mm}
\end{center}
}

\author{
\begin{center}
Lucas~Vittor$^{1,*}$~\orcidlink{0009-0002-6978-0482},
Ayush~Noori$^{1,2,*}$~\orcidlink{0000-0003-1420-1236},
I\~naki~Arango$^{1,*}$~\orcidlink{0009-0002-1443-2325},
Joaquín~Polonuer$^{1}$~\orcidlink{0009-0007-8613-6126}, \\
Sam Rodriques$^{3}$,
Andrew White$^{3}$,
David~A.~Clifton$^{2,4}$~\orcidlink{0000-0002-9848-8555},
Marinka~Zitnik$^{1,5,6,7,\dagger}$~\orcidlink{0000-0001-8530-7228} \\[4mm]
\footnotesize{$^{1}$Department of Biomedical Informatics, Harvard Medical School, Boston, MA, USA} \\
\footnotesize{$^{2}$Department of Engineering Science, University of Oxford, Oxford, UK} \\
\footnotesize{$^{3}$Edison Scientific Inc., San Francisco, CA, USA}\\
\footnotesize{$^{4}$Oxford Suzhou Centre for Advanced Research, University of Oxford, Suzhou, Jiangsu, China}\\
\footnotesize{$^{5}$Broad Institute of MIT and Harvard, Cambridge, MA, USA} \\
\footnotesize{$^{6}$Kempner Institute for the Study of Natural and Artificial Intelligence, Harvard University, MA, USA \!\!\!\!} \\
\footnotesize{$^{7}$Harvard Data Science Initiative, Cambridge, MA, USA} \\[2mm]
\footnotesize{$^*$Co-first authors with equal contribution} \\
\footnotesize{$^{\dagger}$Correspondence: \href{mailto:marinka@hms.harvard.edu}{marinka@hms.harvard.edu}} 
\end{center}
}

\begin{document}
\maketitle

\vspace{1em}
\begin{spacing}{1}
\small
\begin{abstract}
\section*{Abstract}
\input{000abstract}
\end{abstract}
\end{spacing}

\begin{spacing}{1.3}

\section*{Background \& Summary}\label{sec:background}
\input{010background}

\section*{Methods}\label{sec:methods}
\input{020methods}

\section*{Data Record} \label{sec:datarecords}
\input{030datarecords}

\section*{Technical Validation} \label{sec:techvalidation}
\input{040techvalidation}

\section*{Usage Notes} \label{sec:usagenotes}
\input{050usagenotes}

\clearpage
\end{spacing}


\begin{spacing}{1}

\xhdr{Data availability} 
\revision{
\kg is versioned and available through Harvard Dataverse at \url{https://doi.org/10.7910/DVN/IYNGEV}~\autocite{vittor_optimuskg_2026}, under the Creative Commons Attribution-NonCommercial-ShareAlike 4.0 International license (CC BY-NC-SA 4.0). The release includes the complete graph and the largest connected component in Apache Parquet format, together with flat and stratified representations for nodes and edges.
The \kg project provides accompanying documentation at \url{https://optimuskg.ai/docs} and
each release is versioned and linked to the corresponding pipeline code and source definitions used to generate the exported graph. 
}

\xhdr{Code availability}
The \kg codebase, including the data pipeline used to generate the released graph, is released under the MIT License and available at \url{https://github.com/mims-harvard/optimuskg}. \kg data pipeline and analysis dependencies are available at \url{https://github.com/mims-harvard/optimuskg/blob/main/pyproject.toml}. \kg client is available at \url{https://pypi.org/project/optimuskg}.

\xhdr{Acknowledgements}
We thank Vir Mehta for helpful discussions on parts of this manuscript. 

\xhdr{Author contributions}
L.V. led development of the \kg pipeline and core infrastructure, including dataset scope definition, graph structure design, dataset ingestion and transformation, benchmarking of the pipeline core, export functionality, evaluation CLI, and documentation. A.N. defined the initial knowledge graph structure, identified key ontology integration issues, led the PaperQA3 validation analyses. I.A. implemented graph validation and export functionality, resolved ontology integration issues, developed the largest connected component release, created the \kg Python package, and contributed the project website and documentation. J.P. contributed to an early version of the Silver layer of \kg pipeline. S.G.R. and A.W. supported the PaperQA3 analyses. D.A.C. provided feedback on the analyses. L.V., A.N., and M.Z. wrote the manuscript with input from all co-authors. D.A.C. and M.Z. supervised the research, with M.Z. leading the study.

\xhdr{Funding}
We gratefully acknowledge the support of NSF CAREER 2339524, ARPA-H Biomedical Data Fabric (BDF) Toolbox Program, Harvard Data Science Initiative, Amazon Faculty Research, Google Research Scholar Program, AstraZeneca Research, Roche Alliance with Distinguished Scientists (ROADS) Program, Sanofi iDEA-iTECH Award, GlaxoSmithKline Award, Boehringer Ingelheim Award, Merck Award, Optum AI Research Collaboration Award, Pfizer Research, Gates Foundation (INV-079038), Aligning Science Across Parkinson's Initiative (ASAP), Chan Zuckerberg Initiative, John and Virginia Kaneb Fellowship at Harvard Medical School, Biswas Computational Biology Initiative in partnership with the Milken Institute, Collaborative Center for XDP at Massachusetts General Hospital, Harvard Medical School Dean's Innovation Fund for the Use of Artificial Intelligence, and the Kempner Institute for the Study of Natural and Artificial Intelligence at Harvard University. 
A.N. was supported by the Rhodes Scholarship and the Y Combinator Summer Fellows Grant.
D.A.C. was funded by an NIHR Research Professorship (NIHR302440), a Royal Academy of Engineering Research Chair, and the InnoHK Hong Kong Centre for Cerebro-Cardiovascular Engineering, and was supported by the National Institute for Health Research Oxford Biomedical Research Centre and the Pandemic Sciences Institute at the University of Oxford.
Any opinions, findings, conclusions, or recommendations expressed in this material are those of the authors and do not necessarily reflect the views of the funders.

\xhdr{Competing interests}
None.

\end{spacing}

\clearpage

\begin{spacing}{1}
\small
\input{200figures} 
\end{spacing}

\clearpage

\begin{spacing}{1}
\small
\input{300tables} 
\end{spacing}

\clearpage





\section*{References}
\vspace{1em}
\begin{spacing}{1}
\printbibliography[heading=none]
\end{spacing}

\end{document}

%% file: 000abstract.tex
Biomedical knowledge graphs (KGs) are widely used in the life sciences, yet many are derived from unstructured documents and therefore lack schema-level constraints, whereas graphs assembled from structured resources are difficult to harmonize into a unified representation. 
We present \kg, a multimodal biomedical labeled property graph (LPG) built from structured and semi-structured resources to preserve factual, type-specific metadata across molecular, anatomical, clinical, and environmental domains. 
\revision{
\kg contains \numnodes nodes across \numnodetypes entity types, \numedges edges across \numedgetypes edge types, and \numpropertyinstances property instances encoding \numpropertyvalues values across \numpropertykeys distinct property keys, derived from \numontologies ontologies and controlled vocabularies.
} The graph enforces a top-level schema for nodes and edges and retains granular, type-specific properties, cross-references, and provenance.
We assessed the validity of \kg by evaluating whether graph relationships are supported by evidence from the scientific literature using a multimodal agent, PaperQA3. PaperQA3 identified supporting evidence for 70.0\% of sampled edges, whereas 83.4\% of sampled false edges received no supporting evidence. Edges without literature support were concentrated in associations derived from experimental and functional genomics resources, suggesting that \kg captures biomedical knowledge that may precede synthesis in the scientific literature. \kg is distributed as Apache Parquet files, providing a standardized resource for graph-based machine learning, knowledge-grounded retrieval with large language models, and biomedical discovery use cases such as hypothesis generation.

%% file: 010background.tex
AI models that are trained on or that reason over graphs have enabled advances in clinical information extraction~\autocite{nelson_integrating_2019}, disease diagnosis~\autocite{alsentzer_few_2025}, drug discovery~\autocite{cai_pretrainable_2024} and repurposing~\autocite{huang_foundation_2024, zhang_comprehensive_2025}, genetic target identification~\autocite{middleton_phenome-wide_2024}, hypothesis generation~\autocite{noori_graph_2025}, perturbation prediction~\autocite{gonzalez_combinatorial_2025}, spatial tissue analysis~\autocite{ali_graph_2025}, and many other applications across life science and healthcare~\autocite{li_graph_2022, johnson_graph_2024}. Integrating knowledge graphs (KGs) with large language models (LLMs) has been shown to improve generalization~\autocite{tang_graphgpt_2024}, planning~\autocite{wu_can_2024}, reasoning~\autocite{sun_think--graph_2024, tan_paths-over-graph_2025}, and zero-shot learning~\autocite{wang_llms_2024}, including in biomedical settings~\autocite{tian_graph_2024, su_kgarevion_2025}. Critically, the performance of these methods depends on the quality, comprehensiveness, and recency of the underlying graphs~\autocite{luo_learning_2021, vatter_experiment_2025}.

This dependence has catalyzed the development of a vast ecosystem of KGs upon which biomedical graph AI is grounded. 
General purpose biomedical KGs include PrimeKG~\autocite{chandak_building_2023}, BioKG~\autocite{walsh_biokg_2020}, CROssBAR~\autocite{dogan_crossbar_2021}, Hetionet~\autocite{himmelstein_systematic_2017}, MegaKG~\autocite{dong_megakg_2024}, OpenBioLink~\autocite{breit_openbiolink_2020}, PharmKG~\autocite{zheng_pharmkg_2021}, ROBOKOP~\autocite{bizon_robokop_2019}, RTX-KG2~\autocite{wood_rtx-kg2_2022}, and SPOKE~\autocite{morris_scalable_2023}.
Several KGs are designed for drug development and pharmacology, such as BIKG~\autocite{geleta_biological_2021}, DRKG~\autocite{ioannidis_few-shot_2020}, DrugMechDB~\autocite{gonzalez-cavazos_drugmechdb_2023}, IPM-KG~\autocite{tanaka_predicting_2025}, OREGANO~\autocite{boudin_oregano_2023}, NeDRexDB~\autocite{sadegh_network_2021}, and TarKG~\autocite{zhou_tarkg_2024}.
Other KGs have been mined from the literature, like CovidPubGraph~\autocite{pestryakova_covidpubgraph_2022}, GNBR~\autocite{percha_global_2018}, iKraph~\autocite{zhang_comprehensive_2025}, KnowLife~\autocite{ernst_knowlife_2015}, MEDAKA~\autocite{sengupta_medaka_2025}, PKG~\autocite{xu_building_2020}, and SemMedDB~\autocite{kilicoglu_semmeddb_2012}.
Some KGs are centered around specific data modalities; these include CKG~\autocite{santos_knowledge_2022}, GenomicKB~\autocite{feng_genomickb_2023}, \revision{GenomicsKG}~\autocite{jha2019genomicskg}, IDP-KG~\autocite{gray_creating_2022}, ProteinKG65~\autocite{cheng_multi-modal_2022}, and \revision{RNA-KG~\autocite{cavalleri_ontology-based_2024, cavalleri_rna-kg_2026}}.
Still other KGs are specialized to certain diseases, including Alzheimer's disease~\autocite{nian_mining_2022, romano_alzheimers_2024}, colorectal cancer~\autocite{gubanov_cancerkgorg_2024}, COVID-19~\autocite{wang_covid-19_2021, reese_kg-covid-19_2021}, diabetes~\autocite{wang_construction_2020}, hepatocellular carcinoma~\autocite{li_kghc_2020}, Kawasaki disease~\autocite{huang_knowledge_2021}, major depressive disorder~\autocite{huang_constructing_2017}, and rare disease~\autocite{zhu_integrative_2020, alsentzer_few_2025}. 

Despite their widespread use, current biomedical KGs suffer from several limitations~\autocite{bonner_review_2022}. Many KGs do not conform to a standardized schema, creating differences in entity definitions, relation semantics, and identifier harmonization that hinder interoperability. Even widely-used KGs are often released as single snapshots and may rapidly become outdated. The code or data used to construct KGs is often not publicly available or reproducible~\autocite{babalou_reproducible_2023}. Finally, most KGs do not provide structured or free-text metadata about nodes or edges, precluding graph-language learning, nor do they include provenance information, making it difficult to assess evidence quality, update stale assertions, or trace claims back to their original sources. Many KGs also lack built-in quality checks during construction and systematic evaluation procedures for verifying graph elements after release. As a result, existing biomedical KGs often lack consistent schema alignment and provide insufficient provenance and validation, restricting their utility for reproducible data integration and  machine learning applications.

To address these limitations, we developed \kg, a multimodal biomedical knowledge graph that combines ontology-grounded schema alignment with property-rich representations that preserve type-specific metadata. \kg is a labeled property graph (LPG) that adheres to a unified schema and supports granular representations for type-specific node and edge properties. It consists of \numnodes nodes across \numnodetypes entity types (Table~\ref{table:node-types}), \numedges edges across \numedgetypes relation types (Table~\ref{table:edge-types}), and \numpropertyinstances property instances encoding \numpropertyvalues values across \numpropertykeys distinct property keys, derived from heterogeneous datasets (Table~\ref{table:data-sources}) and \numontologies ontologies and controlled vocabularies (Table~\ref{table:ontologies}, Figure~\ref{fig:overview}). \kg integrates entities from multiple ontology trees and has broad ontological coverage across molecular, anatomical, clinical, and environmental domains. \revision{\kg includes environmental exposures in the graph because they connect non-genetic factors to molecular mechanisms and clinical outcomes. \kg therefore links exposures with genes, diseases, biological processes, cellular components, and molecular functions}. To preserve interoperability through standardized identifiers, the upper ontology of \kg is the Biolink Model~\autocite{unni_biolink_2022}. The graph follows the FAIR (Findable, Accessible, Interoperable, and Reusable) principles~\autocite{wilkinson_fair_2016} and includes provenance information to enable evidence tracing and updates. The construction of \kg is enabled by a modular data pipeline that performs ontology harmonization via the BioCypher framework~\autocite{lobentanzer_democratizing_2023}. To assess the validity of \kg, we further evaluated edges in \kg using a literature search and reasoning agent that retrieves scientific evidence and assigns support scores based on the published record~\autocite{white_introducing_2026}. This verification provides an additional quality check on the graph and helps characterize which associations are well supported in the literature and which may originate  from databases or experimental resources. \kg is distributed as Apache Parquet files, providing an efficient format for applications in graph-based machine learning, knowledge-grounded retrieval with large language models, and integrative biomedical analysis.

%% file: 020methods.tex
\subsection*{Development of \kg}\label{sec:development-kg}

We constructed \kg with a reproducible, ontology-grounded data pipeline designed to integrate heterogeneous biomedical resources spanning molecular, anatomical, and clinical scales into a unified labeled property graph. 
The pipeline was fully version-controlled and designed to enable reproducible reconstruction of the knowledge graph from raw data sources.
Data sources were selected based on relevance, curation quality, update frequency, and compatibility with standardized ontologies. Table~\ref{table:data-sources} summarizes the sources present in \kg with their corresponding versions. 

The pipeline followed a medallion data lake architecture~\autocite{medallion2026}, comprising four layers (\textsc{Landing}, \textsc{Bronze}, \textsc{Silver}, \textsc{Gold}) that \revision{organize} data into successive stages of increasing structure, validation, and integration.
The pipeline was orchestrated using the Kedro workflow manager~\autocite{kedro} (Figure~\ref{fig:data-pipeline}). All transformations were defined as modular, deterministic steps within the pipeline, ensuring consistent outputs across executions. At ingestion, the pipeline interfaced with heterogeneous data sources, including APIs (\textit{e.g.}, NCBI), FTP servers (\textit{e.g.}, Open Targets), external databases, static files, and ontology repositories such as BioPortal~\autocite{bioportal}. Access to each resource was mediated through a unified dataset abstraction, in which sources were encapsulated as concrete implementations of a common interface, decoupling source-specific interface heterogeneity from downstream transformations (Table~\ref{table:dataset-abstractions}). Raw data were replicated into the \textsc{Landing} layer as immutable artifacts, preserving original formats (\textit{e.g.}, ZIP, SQL, CSV, JSON, OWL). This design preserved full traceability to original data sources and enabled auditability of all downstream transformations.

The \textsc{Bronze} layer performed initial normalization, converting raw datasets into standardized, machine-readable representations, primarily columnar formats (\textit{i.e.}, Apache Parquet) complemented by semi-structured formats (\textit{i.e.}, JSON). In the \textsc{Silver} layer, the pipeline applied data quality validation (Table~\ref{tab:kedro_hooks}), schema alignment, entity harmonization, and cross-source integration to construct a consistent intermediate graph representation. Edges were first constructed from normalized datasets, after which nodes were materialized from edge endpoints to ensure that all nodes participated in at least one relationship.

The \textsc{Silver} layer constrained all nodes and edges to a shared top-level schema, augmented with type-specific properties. The node and edge schemas were defined in Tables~\ref{table:node-schema} and~\ref{table:edge-schema}, respectively, and all labels were restricted to a predefined set of enumerated values (Table~\ref{table:node-labels}). Schema constraints were enforced at this stage to ensure consistency across entity types and compatibility with the underlying ontology.

The \textsc{Gold} layer produced \kg in Apache Parquet format. Nodes and edges were stored as separate Parquet files, with properties encoded using native columnar types, including nested structures that capture rich scientific metadata (Figure~\ref{fig:metadata}). This representation enabled efficient storage, vectorized processing, and schema-level validation across entity types. The resulting files can be materialized into graph databases (\textit{e.g.}, Neo4j), relational systems (\textit{e.g.}, PostgreSQL), or vector indexes for embedding-based retrieval. These outputs constituted the released dataset and can be directly used or ingested into downstream systems. \revision{Apache Parquet was selected as the supported release format because it embeds a typed schema within each file, and supports the nested, strongly typed columns required by \kg.}

\revision{
\kg follows semantic versioning, with new releases produced when significant changes in the upstream resources occur. Breaking changes to the released graph format or schema are released as major versions (\textit{e.g.}, x.0.0). Non-breaking changes, such as pipeline features, are released as minor versions (\textit{e.g.}, 0.x.0). Documentation changes or fixes that do not affect the data are released as patch versions (\textit{e.g.}, 0.0.x).
}

\revision{
\subsection*{Graph construction}\label{sec:graph-construction}
OptimusKG was the result of a deterministic dataflow rather than a single script. The graph was constructed using a directed acyclic graph (DAG) of 63 deterministic processing nodes that produced 108 materialized, independently inspectable datasets across the \texttt{Landing}, \texttt{Bronze}, \texttt{Silver}, and \texttt{Gold} layers. Each processing node declared its input and output datasets in a central data catalog (Table~\ref{table:dataset-abstractions}), could be executed independently, and wrote its output to disk. Retaining every intermediate dataset was a deliberate design choice to facilitate the diagnosis and audit of the final graph.

Therefore, instead of enumerating 108 different dataset schemas, which are auditable from the pipeline itself, we describe the construction process in terms of its invariants. Two invariants were enforced. First, at most one edge existed between any pair of entities, so that every pair carried a single canonical relationship. Second, every node identifier contained a CURIE whose prefix belonged to the set of namespaces enumerated by the Biolink Model. When a prefix was not available in the primary sources, the pipeline used cross-references as a backup. Entities that could not satisfy the second invariant were excluded from the graph.   

Identifier prefixes (\textit{e.g.}, \texttt{HP:}) were standardized during source-specific normalization, where entities were canonicalized to controlled vocabularies (\textit{e.g.}, using the Human Phenotype Ontology). Records describing the same relation were de-duplicated and consolidated into a single edge with the union of their properties. For cases where the upstream sources disagreed on the relation type, we used a priority map to deterministically select the canonical edge relation (\textit{i.e.}, the \texttt{RELATION\_PRIORITY} defined in the \href{https://github.com/mims-harvard/OptimusKG/blob/af09de36b8f0b61f64eb0f2b137d273b824a675e/optimuskg/pipelines/silver/nodes/constants.py\#L262}{code}), while perserving all source-specific relation assertions in \texttt{properties.relation\_assertions} as \texttt{\{source, relation\}} records. We additionally included a \texttt{properties.relation\_conflict} boolean flag indicating whether the retained assertions contain relations belonging to curated sets of potentially conflicting relation types. Nodes were then constructed from the edge endpoints, so every node participated in at least one relationship. This harmonization process was performed in the \texttt{Silver} layer, with specific sub-packages for the materialization of nodes and edges. Finally, the \texttt{Gold} layer exported the final graph and checked for identifier uniqueness and integrity between edge endpoints. 
}

\subsection*{Ontology harmonization}\label{sec:ontology-harmonization}

While some existing biomedical KGs emphasize scale or breadth, \kg prioritizes ontology-based validation. To ensure consistency between biomedical data sources, the pipeline used the BioCypher framework~\autocite{lobentanzer_democratizing_2023}. \revision{We defined the Biolink Model (v3.2.1)\revision{~\autocite{unni_biolink_2022}} as the upper schema, which provides high-level semantic categories and standardized CURIE namespace conventions. Ontologies} were incorporated through explicit mapping pairs of the form \texttt{(head node, tail node)}, allowing the attachment of additional ontology trees to the base structure and extending semantic granularity. We extended Biolink with the Gene Ontology (GO) \revision{\cite{ashburner2000gene}} and Uber Anatomy Ontology (UBERON) \revision{\cite{mungall2012uberon}} to increase vocabulary resolution across molecular functions (MFN), biological processes (BPO), cellular components (CCO), and anatomical entities (ANA).

Beyond these ontologies, \kg includes entities derived from ontology trees referenced through CURIE namespaces encoded as prefixes of node identifiers. These prefixes correspond to established ontologies (\textit{e.g.}, HPO, DOID, ChEMBL, Orphanet), and therefore the effective ontological coverage of \kg reflects the union of all ontology trees present in the underlying primary data sources. Although entities may originate from external ontologies, BioCypher enforces that all identifiers map to the merged ontology tree. In particular, CURIE prefixes are validated against the set of allowed namespaces specified by Biolink, ensuring that only compliant prefixes are incorporated into the graph. 

This constraint enabled dynamic expansion of ontology coverage in \kg while preserving a consistent semantic representation. Table~\ref{table:ontologies} summarizes the resulting ontological coverage of \kg. \revision{\kg focuses on human biomedical entities and associations. Although some ontologies used for harmonization, such as the Gene Ontology and Uberon, are cross-species resources, \kg retains records derived in a human biomedical context.}

\subsection*{Primary data resources}\label{sec:primary-data-sources}

We implemented dataset-specific content filtering within the \kg data pipeline. Filtering operates across layers and frequently requires joins across intermediate representations. Consequently, filtering decisions are embedded within the broader integration logic rather than executed as isolated, source-specific steps. We summarize here the principal filtering criteria applied to each primary resource, excluding mechanical deduplication (\textit{e.g.}, removal of duplicated entities, repeated elements within lists, redundant property attributes). The access links to the underlying data files for each source at the time of publication can be found in Table~\ref{table:primary-source-access-links}.

\xhdr{Bgee} Bgee provides gene expression data across anatomical and developmental contexts~\autocite{bastian_bgee_2025}. We subset the \textit{Homo sapiens} expression advanced dataset (v2024-05-17) \revision{to} unique Uberon-coded anatomical entities while preserving information on expression rank and call quality. We removed intersection terms (\textit{e.g.}, ``liver $\cap$ cortex'') which represent composite measurements, not discrete tissues.  

\xhdr{Comparative Toxicogenomics Database (CTD)} CTD provides associations between environmental exposures, genes, phenotypes, and diseases~\autocite{davis_comparative_2023}. We filtered exposure data (v2025-10-01) to retain curated human associations and mapped all entities to standardized MeSH, NCBI Gene, and Gene Ontology identifiers, preserving references to associated publications.

\revision{\xhdr{DisGeNET} DisGeNET} is a platform for integrating data on disease-associated genes and variants \autocite{pinero_disgenet_2020}. We separated disease and phenotype data into two entity types.

\xhdr{DrugBank} DrugBank is a database of drugs and drug targets, along with chemical, pharmacological, and pharmaceutical metadata~\autocite{knox2024drugbank}. We extracted drug-target, drug-enzyme, drug-carrier, and drug-transporter associations, mapping protein identifiers to standardized gene identifiers.

\xhdr{DrugCentral} DrugCentral is a drug information resource that provides information on active ingredients, mechanisms of action, pharmaceutical approvals, and other drug properties ~\autocite{avram_drugcentral_2023}. We included only drug entities where both CAS registry number and UMLS CUI could be cross-referenced. We divided drug-disease and drug-phenotype edges.

\xhdr{Gene names} Standardized gene names and symbols for all human genes, including protein-coding genes, non-coding RNA, and pseudogenes, were retrieved from the HUGO Gene Nomenclature Committee (HGNC)~\autocite{seal2025genenames}.

\xhdr{OnSIDES} OnSIDES is a database of adverse drug events extracted from FDA Structured Product Labels by a fine-tuned PubMedBERT language model~\autocite{tanaka_onsides_2025}. We included only drug identifiers with \revision{purely} numeric (\texttt{\textasciicircum\textbackslash d+\$}) RxNorm concepts.

\xhdr{Open Targets} The Open Targets Platform aggregates gene target-disease and drug target associations~\autocite{buniello_open_2025}. In addition to drugs and targets, this database also contains information about diseases, phenotypes, variants, and genome-wide association studies (GWAS). We deduplicated records that have the same name but different ontology identifiers. For cross-reference mappings, we required that the database prefix starts with UMLS identifiers for downstream disease harmonization. 

\xhdr{Protein-protein interactions} Human protein-protein interaction information was included from \textcite{menche_uncovering_2015}, BioGRID \autocite{oughtred_biogrid_2021}, STRING \autocite{szklarczyk_string_2025}, and the Human Reference Interactome generated by \textcite{luck_reference_2020}. We removed duplicate protein-protein interaction edges.

\xhdr{Reactome} Reactome is a database of molecular reactions, interactions, and biological pathways curated from the scientific literature~\autocite{ragueneau_reactome_2026}. We included only human pathways.

Because \kg integrates resources with varying scope and coverage, the distribution of edges across relation types reflects the composition of underlying data sources. In particular, large-scale resources such as Open Targets contribute a substantial fraction of associations, which may introduce imbalances in coverage across entity and relation types.

\subsection*{Primary ontologies}\label{sec:ontologies}

\xhdr{Biolink} The Biolink Model is a data model that provides a standardized vocabulary for biomedical entities, including genes, diseases, chemicals, anatomical structures, and phenotypes, across knowledge graphs~\autocite{unni_biolink_2022}.

\xhdr{Human Disease Ontology Knowledgebase (DO-KB)} DO-KB is a classification of genetic, infectious, cancer, environmental, complex, rare, and common human diseases represented in the semantic Web Ontology Language (OWL)~\autocite{baron_-kb_2026}. DO-KB includes disease features and relationships along with cross-references to other clinical vocabularies.

\xhdr{Gene Ontology (GO)} GO is an ontology that describes the functions of genes in three categories: molecular functions, cellular components, and biological processes~\autocite{ashburner_gene_2000, the_gene_ontology_consortium_gene_2026}.

\xhdr{Human Phenotype Ontology (HPO)} HPO is a vocabulary of the phenotypic features of human disease~\autocite{gargano_human_2024}.

\xhdr{Mondo Disease Ontology} Mondo is a disease classification framework that harmonizes disease definitions in Online Mendelian Inheritance in Man (OMIM), Orphanet, Medical Subject Headings (MeSH), National Cancer Institute Thesaurus (NCIt) and other databases~\autocite{vasilevsky_mondo_2025}.

\xhdr{Orphanet Rare Disease Ontology (ORDO)} ORDO is a structured vocabulary for rare diseases, associated genes, and clinical features derived from the Orphanet database~\autocite{weinreich_orphanet_2008}. 

\xhdr{Uber-anatomy ontology (Uberon)} Uberon is a cross-species anatomical ontology that contains information about anatomical, phenotypic, and expression data, along with anatomical entities classified by structure, function, and developmental lineage~\autocite{mungall_uberon_2012}.

As with any integrated resource, \kg inherits limitations from its source datasets, including potential inconsistencies in curation quality, coverage, and update frequency.

\revision{The version of \kg released with this paper encodes \numnodetypes node types spanning molecular, clinical, anatomical, and environmental entities (Figure~\ref{fig:overview}A, Table~\ref{table:node-types}), as well as \numedgetypes edge types spanning molecular, clinical, anatomical, and environmental interactions (Figure~\ref{fig:overview}B, Table~\ref{table:edge-types}). The cardinalities of connected components in \kg are summarized in Table~\ref{table:connected-components}. The \kg pipeline (Figure~\ref{fig:data-pipeline}) supports periodic updates as underlying data sources evolve, enabling regeneration of the graph with updated versions of primary datasets and ontologies. The download links for the ontologies used to create the graph at the time of publication can be found in Table~\ref{table:ontology-download-links}.}

%% file: 030datarecords.tex
\subsection*{Release documents}\label{sec:release-documents} 

\revision{\kg is available via Harvard Dataverse at \url{https://doi.org/10.7910/DVN/IYNGEV} ~\autocite{vittor_optimuskg_2026}.} \kg is released in Apache Parquet format in two distributions: the complete graph and its largest connected component. The complete release provides the full graph as separate node and edge tables, whereas the second distribution contains the largest connected component (LCC) of \kg. Because node and edge types have heterogeneous, type-specific schemas (Figure~\ref{fig:metadata}a), the flat \texttt{nodes.parquet} and \texttt{edges.parquet} files serialize properties as JSON strings. To support native typed access, \kg also provides stratified Parquet files in the \texttt{nodes/} and \texttt{edges/} subdirectories, with one file per node or edge type, in which the \texttt{properties} field is materialized as a nested \texttt{Struct} with native Polars types (\eg, \texttt{nodes/gene.parquet}, \texttt{edges/anatomy\_gene.parquet}) (Figure~\ref{fig:metadata}b).

\subsection*{Node records}\label{sec:node-records}

All nodes follow a standardized base schema comprising \code{id}[String] (unique identifier) and \code{label}[String] (\textit{e.g.}, GEN). Provenance is tracked via \code{sources}[\{direct:String[], indirect:String[]\}]. Each node includes a \code{properties}[struct] key containing type-specific attributes. The following sections describe the node types in \kg and their corresponding properties. 

\xhdr{Gene (GEN)} \revision{Because most upstream sources mix genes, canonical proteins, isoforms, and other gene products, protein annotations are represented as gene properties.} \code{symbol}[String] (approved gene symbol), \code{name}[String] (human-readable gene name), and \code{biotype}[String] (gene biotype classification) record core gene identity. Transcript-level attributes include \code{transcript\_ids}[String[]] (list of transcript identifiers), \code{canonical\_transcript}[\{id:String, chromosome:String, start:Int64, end:Int64, strand:String\}] (primary transcript with identifier, genomic coordinates, and strand), and \code{canonical\_exons}[String[]] (identifiers of canonical exons), while genomic coordinates are stored in \code{genomic\_location}[\{chromosome:String, start:Int64, end:Int64, strand:Int32\}] (genomic locus defined by chromosome, start/end positions, and strand). Synonym and cross-reference resolution is encoded via \code{synonyms}[\{label:String, source:String\}[]] (gene synonyms with label and provenance source), \code{symbol\_synonyms}[\{label:String, source:String\}[]] (alternative gene symbols with source), \code{name\_synonyms}[\{label:String, source:String\}[]] (alternative gene names with source), and \code{xrefs}[\{id:String, source:String\}[]] (external database cross-references with identifier and source), with deprecated labels via \code{obsolete\_symbols}[\{label:String, source:String\}[]] (deprecated symbols with source) and \code{obsolete\_names}[\{label:String, source:String\}[]] (deprecated names with source). Functional characterization includes \code{function\_descriptions}[String[]] (free-text functional annotations), \code{subcellular\_locations}[\{location:String, source:String, term\_sl:String, label\_sl:String\}[]] (subcellular localization with location, ontology term, label, and source), \code{target\_class}[\{id:Int64, label:String, level:String\}[]] (drug target classification with identifier, label, and hierarchy level), \code{hallmarks\_attributes}[\{pmid:Int64, description:String, attribute\_name:String\}[]] (cancer hallmark attributes with literature reference, description, and attribute name), and \code{cancer\_hallmarks}[\{pmid:Int64, description:String, impact:String, label:String\}[]] (cancer hallmark classifications with reference, description, impact, and label). Relational context is provided by \code{homologues}[\{species\_id:String, species\_name:String, homology\_type:String, target\_gene\_id:String, is\_high\_confidence:String, target\_gene\_symbol:String, query\_percentage\_identity:float, target\_percentage\_identity:float, priority:Int64\}[]] (homologous genes with species metadata, target identifiers, homology type, confidence, sequence identity metrics, and priority) and \code{alternative\_genes}[String[]] (alternative gene identifiers). Drug discovery fields include \code{tractability}[\{modality:String, id:String, value:bool\}[]] (drug tractability assessments with modality, identifier, and boolean value), \code{safety\_liabilities}[\{event:String, event\_id:String, effects:\{direction:String, dosing:String\}[], biosamples:\{cell\_format:String, cell\_label:String, tissue\_id:String, tissue\_label:String\}[], datasource:String, literature:String, url:String, studies:\{description:String, name:String, type:String\}[]\}[]] (safety liabilities with event metadata, nested effects and biosamples, provenance, references, and supporting studies), \code{target\_enabling\_package}[\{target\_from\_source\_id:String, description:String, therapeutic\_area:String, url:String\}] (target enabling package with source identifier, description, therapeutic area, and reference URL), \code{chemical\_probes}[\{control:String, drug\_id:String, id:String, is\_high\_quality:bool, mechanism\_of\_action:String[], origin:String[], probe\_miner\_score:Int64, probes\_drugs\_score:Int64, score\_in\_cells:Int64, score\_in\_organisms:Int64, target\_from\_source\_id:String, urls:\{nice\_name:String, url:String\}[]\}[]] (chemical probes with identifiers, quality flags, mechanisms, origin, multiple scoring metrics, target mapping, and external URLs), and \code{associated\_proteins}[\{id:String, source:String\}[]] (linked protein identifiers with source). Constraint scores are in \code{constraint\_scores}[\{constraint\_type:String, score:float, exp:float, obs:Int64, oe:float, oe\_lower:float, oe\_upper:float, upper\_rank:Int64, upper\_bin:Int64, upper\_bin6:Int64\}[]] (constraint and intolerance metrics including expected/observed counts, ratios, confidence bounds, and ranking bins) and transcriptional context in \code{transcription\_start\_site}[Int64] (TSS coordinate).

\xhdr{Drug (DRG)} Represents pharmacological compounds with chemical, regulatory, and clinical information. Identity fields include \code{name}[String] (drug name), \code{inchi\_key}[String] (InChI key identifier), \code{type}[String] (drug classification), and \code{canonical\_smiles}[String] (canonical SMILES representation). Chemical attributes are captured through \code{cd\_formula}[String] (molecular formula), \code{cd\_mol\_weight}[float] (molecular weight), \code{calculated\_log\_p}[float] (lipophilicity estimate), \code{alogs}[float] (predicted aqueous solubility descriptor), \code{tpsa}[float] (topological polar surface area), \code{lipinski}[float] (Lipinski rule violations score), and structural descriptors including \code{aromatic\_carbons}[Int32] (count of aromatic carbons), \code{sp3\_count}[Int32] (count of sp\textsuperscript{3}-hybridized atoms), \code{sp2\_count}[Int32] (count of sp\textsuperscript{2}-hybridized atoms), \code{sp\_count}[Int32] (count of sp-hybridized atoms), \code{halogen\_count}[Int32] (count of halogen atoms), \code{hetero\_sp2\_count}[Int32] (count of sp\textsuperscript{2} heteroatoms), \code{rotatable\_bonds}[Int32] (count of rotatable bonds), \code{o\_n}[Int32] (count of oxygen and nitrogen atoms), and \code{oh\_nh}[Int32] (count of OH and NH groups). Regulatory and clinical status are encoded in \code{is\_approved}[bool] (approval status flag), \code{has\_been\_withdrawn}[bool] (withdrawal status flag), \code{maximum\_clinical\_trial\_phase}[float] (maximum clinical trial phase reached), \code{year\_of\_first\_approval}[Int64] (year of first approval), \code{black\_box\_warning}[bool] (presence of FDA black box warning), and \code{status}[String] (drug development or regulatory status), with additional context provided by \code{fda\_labels}[Int32] (count of FDA labels) and \code{number\_of\_formulations}[Int32] (count of available formulations). Naming and cross-references are represented via \code{synonyms}[String[]] (alternative drug names), \code{trade\_names}[String[]] (brand or trade names), \code{accession\_numbers}[String[]] (accession identifiers), \code{sources}[\{direct:String[], indirect:String[]\}] (data provenance with direct and indirect source lists), \code{source\_ids}[String[]] (source-specific identifiers), \code{chemical\_abstracts\_service\_number}[String] (CAS registry number), and \code{unique\_ingredient\_identifier}[String] (UNII identifier). Structural representations are provided via \code{mol\_file\_base64}[String] (base64-encoded MOL file), \code{mol\_image\_base64}[String] (base64-encoded molecular structure image), \code{inchi}[String] (InChI string), \code{mrdef}[String] (molecular representation description), \code{enhanced\_stereo}[bool] (enhanced stereochemistry flag), \code{cd\_id}[String] (calculated descriptor identifier), \code{struct\_id}[String] (structure identifier), and \code{rgb}[float] (numeric RGB encoding). Free-text annotation is captured in \code{description}[String] (textual description of the drug).

\xhdr{Disease (DIS)} Encodes human diseases with hierarchical ontology structure and clinical annotations. Identity is resolved by \code{name}[String] (disease name), \code{description}[String] (free-text disease description), and \code{code}[String] (disease identifier). \revision{Ontological relationships are represented as edges.} Synonymy spans \code{exact\_synonyms}[String[]] (exact lexical synonym matches), \code{related\_synonyms}[String[]] (semantically related synonyms), \code{broad\_synonyms}[String[]] (broader terms), and \code{narrow\_synonyms}[String[]] (narrower terms), with deprecated terminology tracked in \code{obsolete\_terms}[String[]] (deprecated disease terms) and \code{obsolete\_xrefs}[String[]] (deprecated external references). Cross-database reference is enabled through \code{xrefs}[String[]] (external cross-references), \code{concept\_ids}[String[]] (linked concept identifiers), \code{concept\_names}[String[]] (linked concept names), \code{umls\_cui}[String] (UMLS Concept Unique Identifier), \code{snomed\_full\_names}[String[]] (SNOMED CT preferred names), and \code{snomed\_concept\_ids}[String[]] (SNOMED CT concept identifiers). Clinical context is captured via \code{therapeutic\_areas}[String[]] (therapeutic area classifications) and \code{cui\_semantic\_type}[String] (UMLS semantic type classification).

\xhdr{Phenotype (PHE)} Represents clinical features and phenotypic abnormalities. Identity attributes include \code{name}[String] (phenotype name), \code{description}[String] (free-text phenotype description), and \code{code}[String] (phenotype identifier). \revision{Hierarchical relationships are represented as edges.} Synonym resolution is encoded via \code{exact\_synonyms}[String[]] (exact lexical synonym matches), \code{related\_synonyms}[String[]] (semantically related synonyms), \code{broad\_synonyms}[String[]] (broader terms), and \code{narrow\_synonyms}[String[]] (narrower terms), with deprecated terminology tracked in \code{obsolete\_terms}[String[]] (deprecated phenotype terms) and \code{obsolete\_xrefs}[String[]] (deprecated external references). Ontology metadata is retained through \code{ontology}[\{title:String, version:String, license:String, description:String\}] (ontology metadata including title, version, license, and description), and \code{type}[String] (phenotype classification or ontology type). Clinical mappings are provided via \code{concept\_ids}[String[]] (linked concept identifiers), \code{concept\_names}[String[]] (linked concept names), \code{umls\_cui}[String] (UMLS Concept Unique Identifier), \code{snomed\_full\_names}[String[]] (SNOMED CT preferred names), \code{snomed\_concept\_ids}[String[]] (SNOMED CT concept identifiers), and \code{cui\_semantic\_type}[String] (UMLS semantic type classification).

\xhdr{Anatomy (ANA)} Encodes anatomical structures and tissues. Identity descriptors include \code{name}[String] (anatomical structure name) and \code{definition}[String] (free-text anatomical definition). Cross-references and synonyms are captured via \code{xrefs}[String[]] (external ontology and database cross-references) and \code{synonyms}[String[]] (alternative anatomical names and labels). Ontology metadata is preserved through \code{ontology}[\{title:String, version:String, license:String, description:String\}] (ontology metadata including title, version, license, and description).

\xhdr{Pathway (PWY)} Represents biological pathways capturing molecular interactions. Identity fields include \code{name}[String] (pathway name) and \code{species}[String] (organism or taxonomic label associated with the pathway).

\xhdr{Biological process (BPO)} Encodes Gene Ontology biological processes describing coordinated biological activities. Identity descriptors include \code{name}[String] (biological process name) and \code{definition}[String] (formal ontology definition of the process). Cross-references and synonymy are represented via \code{xrefs}[String[]] (external ontology and database cross-references) and \code{synonyms}[String[]] (alternative process names and labels). Ontology metadata is encoded via \code{ontology}[\{title:String, version:String, license:String, description:String\}] (ontology metadata including title, version, license, and description). Provenance is captured through \code{sources}[\{direct:String[], indirect:String[]\}] (data provenance with direct and indirect source lists).

\xhdr{Cellular component (CCO)} Represents Gene Ontology cellular components corresponding to subcellular structures and compartments. Mirrors biological process (BPO) schema. 

\xhdr{Molecular function (MFN)} Encodes Gene Ontology molecular functions describing biochemical and enzymatic activities. Mirrors biological process (BPO) schema. 

\xhdr{Exposure (EXP)} Captures chemical, physical, and environmental exposures linked to biological and clinical outcomes. Identity is resolved via \code{name}[String] (exposure name), \code{source\_categories}[String[]] (categories describing the origin or context of the exposure), and \code{source\_details}[String] (free-text description providing additional source-specific details).

\subsection*{Edge records}\label{sec:edge-types}

All edges follow a standardized base schema comprising \code{from}, \code{to} (node identifiers), \code{label} (\textit{e.g.}, ANA-GEN), \code{relation} (relationship type), \code{undirected} (directionality flag). Provenance is captured via \code{sources}[\{direct:String[], indirect:String[]\}]. In addition, each edge includes a \code{properties}[struct] key containing type-specific attributes. Edges retain source-specific relation information through \texttt{relation\_assertions:\{source:String, relation:String\}[]}, which records source-relation assertions consolidated into the edge, and \texttt{relation\_conflict:Boolean}, which flags potential conflicts among those assertions. The following sections describe the edge types in \kg and their corresponding properties. 

\xhdr{Anatomy-Gene (ANA-GEN)} Links anatomical structures to genes expressed in those tissues. Quantitative expression is captured via \code{expression\_rank}[Int32] (relative expression ranking across tissues), and confidence through \code{call\_quality}[String] (qualitative confidence level of the expression call).

\xhdr{Anatomy-Anatomy (ANA-ANA)} Represents hierarchical relationships (\textit{e.g.}, part-of, adjacent-to) between anatomical structures.

\xhdr{Disease-Gene (DIS-GEN)} Encodes disease-gene associations. Evidence strength is captured through \code{evidence\_score}[float] (aggregate evidence score), \code{evidence\_count}[Int64] (number of supporting evidence records), \code{evidence\_index}[float] (normalized evidence index), and \code{disgenet\_score}[float] (DisGeNET association score), with specificity and pleiotropy measured by \code{disease\_specificity\_index}[float] (degree of disease specificity) and \code{disease\_pleiotropy\_index}[float] (degree of gene pleiotropy across diseases). Temporal coverage is encoded via \code{year\_initial}[String] (earliest reported association year) and \code{year\_final}[String] (most recent reported association year), with supporting evidence quantified through \code{number\_of\_pmids}[Int32] (number of supporting publications) and \code{number\_of\_snps}[Int32] (number of associated SNPs).

\xhdr{Disease-Disease (DIS-DIS)} Represents hierarchical (\textit{e.g.}, parent) and semantic (\textit{e.g.}, synonym) relationships between diseases.

\xhdr{Disease-Phenotype (DIS-PHE)} Links diseases to phenotypic features. Clinical descriptors include \code{aspect}[String[]] (phenotype aspect categories), \code{evidence\_type}[String[]] (evidence codes supporting the association), \code{frequency}[String[]] (phenotype frequency annotations or ontology terms), \code{onset}[String[]] (age of onset categories), \code{modifiers}[String[]] (phenotypic modifiers), and \code{sexes}[String[]] (sex-specific occurrence annotations), with negation captured through \code{qualifier\_not}[bool] (flag indicating explicit absence of the phenotype). Bibliographic support and curation are recorded in \code{bio\_curation}[String[]] (curation provenance strings with curator and date) and \code{references}[String[]] (supporting literature and database references).

\xhdr{Phenotype-Gene (PHE-GEN)} Encodes associations between genes and phenotypic features. Properties mirror disease-gene (DIS-GEN) associations.

\xhdr{Phenotype-Phenotype (PHE-PHE)} Represents hierarchical relationships between phenotypic features, including compositional (\textit{e.g.}, is\_a) and part-whole (\textit{e.g.}, has\_part) structures.

\xhdr{Drug-Gene (DRG-GEN)} Captures drug-target interactions with pharmacological annotations. Mechanistic details are encoded via \code{mechanisms\_of\_action}[String[]] (descriptions of the biochemical or pharmacological mechanisms of action), with cross-references captured in \code{source\_ids}[String[]] (source-specific identifiers) and \code{source\_urls}[String[]] (external URLs linking to supporting resources).

\xhdr{Drug–Disease (DRG–DIS)} Encodes drug–disease associations including therapeutic and clinical development context. Clinical progression is represented through \code{highest\_clinical\_trial\_phase}[float] (maximum clinical trial phase achieved for the drug–disease association), with supporting identifiers in \code{structure\_id}[String] (internal or source-specific structure identifier), \code{drug\_disease\_id}[String] (association identifier), and \code{reference\_ids}[String[]] (external reference identifiers supporting the association).

\xhdr{Drug–Phenotype (DRG–PHE)} Represents drug effects on phenotypic outcomes. Mirrors drug-disease (DRG-DIS) schema.

\revision{\xhdr{Drug-Biological process (DRG-BPO)} Encodes drug indications whose therapeutic target is a physiological process rather than a disease. Clinical trial information is preserved via  \texttt{reference\_ids:String[]}, with development status via \texttt{highest\_\allowbreak clinical\_\allowbreak trial\_\allowbreak phase:Float}.}

\xhdr{Drug–Drug (DRG–DRG)} Encodes drug–drug interactions, including combinatorial and adverse interaction effects via \code{interaction\_description}[String].

\xhdr{Gene–Gene (GEN–GEN)} Represents gene–gene (\textit{e.g.}, protein-protein, genetic) interactions.

\xhdr{Biological process–Gene (BPO–GEN)} Encodes gene participation in biological processes. Gene Ontology evidence codes are captured via \code{evidence}[String[]] (GO evidence codes supporting the association), with gene product context provided via \code{gene\_product}[String[]] (identifiers of gene products associated with the process) and \code{eco\_ids}[String[]] (Evidence and Conclusion Ontology identifiers corresponding to the evidence).

\xhdr{Molecular function–Gene (MFN–GEN)} Encodes gene products and their molecular functions. Mirrors biological process-gene (BPO-GEN) schema.

\xhdr{Cellular component–Gene (CCO–GEN)} Encodes gene localization to cellular components. Mirrors biological process-gene (BPO-GEN) schema.

\xhdr{Biological process–Biological process (BPO–BPO)} Represents hierarchical relationships between biological processes, including part–whole (\textit{e.g.}, part-of) and temporal ordering (\textit{e.g.}, precedes) relationships.

\xhdr{Cellular component–Cellular component (CCO–CCO)} Represents hierarchical (\textit{e.g.}, part-of) and spatial relationships (\textit{e.g.}, located\_in) between cellular components.

\xhdr{Molecular function–Molecular function (MFN–MFN)} Represents hierarchical relationships (\textit{e.g.}, part\_of, related\_to) between molecular functions.

\xhdr{Pathway–Gene (PWY–GEN)} Encodes gene participation in biological pathways, linking molecular entities to higher-level functional modules.

\xhdr{Pathway–Pathway (PWY–PWY)} Represents hierarchical (\textit{e.g.}, part\_of) and compositional relationships (\textit{e.g.}, related\_to) between pathways.

\xhdr{Exposure–Gene (EXP–GEN)} Encodes exposure–gene interactions with epidemiological context. Evidence is recorded through \code{evidence\_count}[Int32] (number of supporting evidence records). Receptor context is provided via \code{number\_of\_receptors}[Int32] (count of associated receptors), \code{receptors}[String[]] (receptor identifiers or names), and \code{receptor\_notes}[String[]] (free-text receptor annotations). Demographic stratification includes \code{smoking\_statuses}[String[]] (smoking status categories), \code{sexes}[String[]] (sex-specific annotations), \code{races}[String[]] (reported race or ethnicity categories), and age distributions via \code{age\_entries}[String[]] (age category entries), \code{age\_range\_values}[String[]] (age ranges), \code{age\_mean\_values}[String[]] (mean age values), \code{age\_median\_values}[String[]] (median age values), \code{age\_point\_values}[String[]] (point estimates of age), and \code{age\_open\_range\_values}[String[]] (open-ended age ranges). Experimental and analytical context is captured via \code{methods}[String[]] (experimental or analytical methods), \code{detection\_limit}[String[]] (detection threshold values), \code{detection\_limit\_uom}[String[]] (units of measurement for detection limits), \code{detection\_frequency}[String[]] (frequency of detection), \code{mediums}[String[]] (biological or environmental sample types), and \code{assay\_notes}[String[]] (free-text assay descriptions). Geographic metadata includes \code{study\_countries}[String[]] (countries of study), \code{states\_or\_provinces}[String[]] (subnational regions), and \code{city\_town\_region\_areas}[String[]] (local geographic areas). Outcome and study context are encoded via \code{outcome\_relationships}[String[]] (type of exposure–outcome relationship), \code{exposure\_outcome\_notes}[String[]] (free-text outcome annotations), \code{references}[String[]] (supporting literature references), \code{associated\_study\_titles}[String[]] (titles of associated studies), \code{enrollment\_start\_years}[String[]] (study enrollment start years), \code{enrollment\_end\_years}[String[]] (study enrollment end years), \code{study\_factors}[String[]] (study design or covariates), and \code{exposure\_event\_notes}[String[]] (free-text exposure event descriptions).

\xhdr{Exposure–Disease (EXP–DIS)} Encodes environmental exposure–disease associations. Mirrors exposure-gene (EXP-GEN) schema.

\xhdr{Exposure–Exposure (EXP–EXP)} Represents compositional and hierarchical relationships between exposure types. Mirrors exposure-gene (EXP-GEN) schema.

\xhdr{Exposure–Biological process (EXP–BPO)} Encodes environmental exposure effects on biological processes. Mirrors exposure-gene (EXP-GEN) schema.

\xhdr{Exposure–Cellular component (EXP–CCO)} Encodes environmental exposure effects on cellular components. Mirrors exposure-gene (EXP-GEN) schema.

\xhdr{Exposure–Molecular function (EXP–MFN)} Encodes environmental exposure effects on molecular functions. Mirrors exposure-gene (EXP-GEN) schema.

%% file: 040techvalidation.tex
To assess the biological and clinical validity of edges in \kg as a data resource, we evaluated whether graph relationships are supported by evidence from the scientific literature.
We used PaperQA3, a multimodal AI agent for deep research~\autocite{lala_paperqa_2023, white_introducing_2026} that synthesizes evidence from scientific literature. PaperQA3 retrieves information from over 150 million research papers and patents and aggregates supporting evidence for scientific queries. It has been shown to match human expert performance on literature review and search~\autocite{skarlinski_language_2024} and we use it here to assess whether edges in \kg are supported by evidence from the scientific literature~\autocite{noori_graph_2025}.

We first performed stratified sampling of seed nodes across all node types represented in \kg. The top and bottom 10\% of nodes by within-type undirected degree centrality were excluded from sampling to avoid concepts that are extremely generic hubs or peripheral nodes with insufficient connectivity for evaluation, respectively (\revision{Figure~\ref{fig:topology}}). From the remaining pool, we uniformly sampled 100 seed nodes per type, for a total of 1,000 seed nodes. For each seed, we constructed a validation set of true edges by sampling up to 10 distinct neighbors in \kg with non-empty descriptions. 743 seed nodes had degree $< 10$; for these nodes, the full neighborhood was sampled for evaluation. As a negative control, we also sampled one false neighbor per seed not connected to the seed in \kg. The final evaluation set consisted of 459 edges incident on exposure nodes (EXP, $\bar{d} = 5.20$, where $\bar{d}$ represents the mean degree across sampled seeds, Figure~\ref{fig:paperqa}a), 343 edges incident on cellular component nodes (CCO, $\bar{d} = 3.85$, Figure~\ref{fig:paperqa}b), 164 edges incident on anatomy nodes (ANA, $\bar{d} = 1.64$, Figure~\ref{fig:paperqa}c), 274 edges incident on molecular function nodes (MFN, $\bar{d} = 2.74$, Figure~\ref{fig:paperqa}d), 690 edges incident on pathway nodes (PWY, $\bar{d} = 12.37$, Figure~\ref{fig:paperqa}e), 367 edges incident on phenotype nodes (PHE, $\bar{d} = 4.03$, Figure~\ref{fig:paperqa}f), 444 edges incident on biological process nodes (BPO, $\bar{d} = 4.70$, Figure~\ref{fig:paperqa}g), 439 edges incident on drug nodes (DRG, $\bar{d} = 58.68$, Figure~\ref{fig:paperqa}h), 528 edges incident on disease nodes (DIS, $\bar{d} = 37.25$, Figure~\ref{fig:paperqa}i), and 1,000 edges incident on gene nodes (GEN, $\bar{d} = 229.27$, Figure~\ref{fig:paperqa}j), for a total of 4,708 true edges, along with 1,000 false edges. 

For each edge, PaperQA3 was prompted to retrieve relevant literature and assign an integer evidence score on a scale from $1$ to $5$, where $1$ indicates no supporting evidence found, $2$ indicates weak evidence, with 1-2 papers in support; $3$ indicates moderate evidence, with 3-4 papers in support; $4$ indicates strong evidence, with 5-6 papers in support; and $5$ indicates very strong evidence, with over 6 supporting papers or substantial experimental evidence. 
PaperQA3 also produced a rationale for each evaluation along with references to supporting peer-reviewed scientific or clinical evidence. The agent generated a successful response for 5,706 of the 5,708 queries. This procedure provides a quantitative measure of literature support for graph edges and enables assessment of \kg dataset validity.

Literature evidence was identified for 70.0\% of sampled edges in \kg ($n = 3,293$); of those, 12.2\% were weakly supported (\ie, 1-2 papers, $n = 575$), 17.6\% were moderately supported (\ie, 3-4 papers, $n = 826$), 10.9\% were strongly supported (\ie, 5-6 papers, $n = 511$), and 29.3\% were very strongly supported (\ie, greater than 6 papers, $n = 1,381$) (Figure~\ref{fig:paperqa}). PaperQA3 could not find evidence to support 30\% of edges ($n = 1,413$); of these, the most prevalent edge type was ANA-GEN ($n = 661$, 46.8\% of rating $= 1$, Figure~\ref{fig:paperqa}j), followed by DIS-GEN ($n = 204$, 14.4\% of rating $= 1$, Figure~\ref{fig:paperqa}j) and DRG-DRG ($n = 118$, 8.4\% of rating $= 1$, Figure~\ref{fig:paperqa}h). This pattern likely reflects a lag between primary experimental data deposition and its synthesis in the scientific literature. Many gene expression, disease target, and drug-drug interaction edges in \kg were derived from primary experimental data or functional genomics screens available via Open Targets~\autocite{buniello_open_2025} or similar platforms. Associations are often deposited in these platforms as raw data before they are synthesized into narrative text, and as such, would not be accessible to literature search agents like PaperQA3. By contrast, 83.4\% ($n = 834$) of false edges were not supported by any evidence, and only 1.4\% ($n = 14$) of false edges were very strongly supported, indicating that the graph has high specificity with respect to literature-supported relationships.
Together, these results indicate that \kg is broadly consistent with the scientific literature and also includes experimental data that is not yet widely represented in published studies. These analyses demonstrate a scalable agentic approach for assessing the validity of large knowledge graphs using literature evidence.

%% file: 050usagenotes.tex
 \revision{

\kg is intended for biomedical research and scientific AI, providing a standardized resource for graph-based machine learning, knowledge-grounded retrieval with large language models, and biomedical discovery use cases such as hypothesis generation.

The supported release format is Apache Parquet, which can be loaded through the \texttt{optimuskg} Python client. \kg can be used as a columnar DataFrame or NetworkX MultiDiGraph, with the option to point to specific node and edge types and the largest connected component of the graph. Additional community-developed integrations are listed in the project website.

For use with AI agents, \kg provides an agent skill that describes how to load, query, filter, and analyze the graph. \kg can also be used via the Model Context Protocol Bundle (MCPB) file, which provides natural-language access through graph-specific tools backed by a locally indexed database.
}

%% file: 200figures.tex
\begin{figure}[!t]
  \centering
  \includegraphics[width=\linewidth]{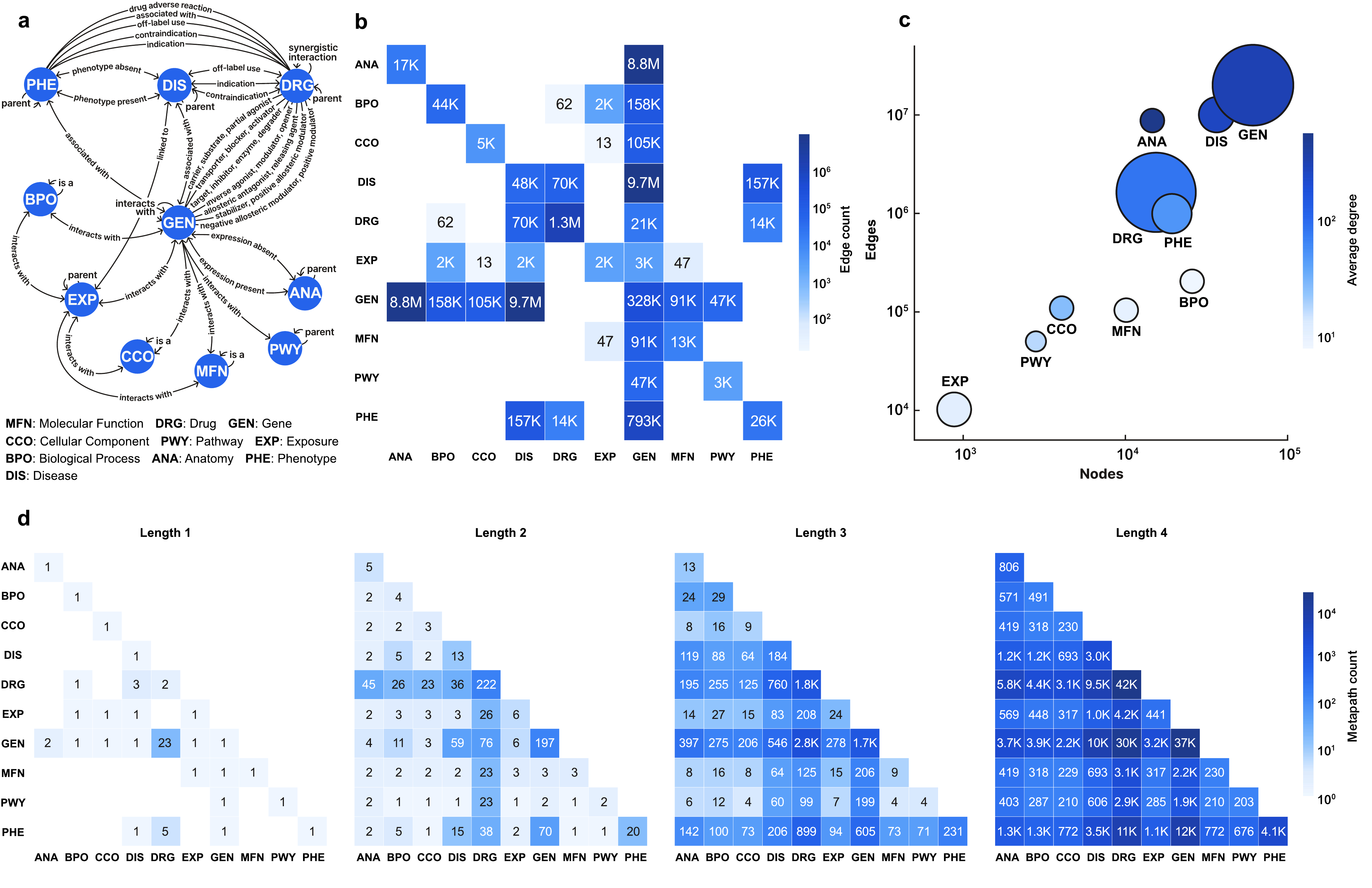}
  \caption{\revision{\textsbf{Overview of \kg.} 
  \textsbf{(a)} Metagraph of \kg, illustrating the node types and the heterogeneous relationships connecting them. \textsbf{(b)} Pairwise edge type distribution across the graph. Each cell reports the number of edges between two node types. Edges remain asymmetric; a reverse edge is added only when the \code{undirect} key of the edge is set to \code{True}. \textsbf{(c)} Total number of nodes ($x$-axis) and edges ($y$-axis) for each entity type. Bubble size is proportional to the number of non-null properties associated with the entities. \textsbf{(d)} Metapath counts across increasing path lengths (1-4). Metapaths are defined as unique triples of the form (\code{from}, \code{relation}, \code{to}). Only the lower triangle is stored, as all edges are treated as bidirectional.
  }}
  \label{fig:overview}
\end{figure}
\clearpage

\begin{figure}[!t]
  \centering
  \includegraphics[width=1\linewidth]{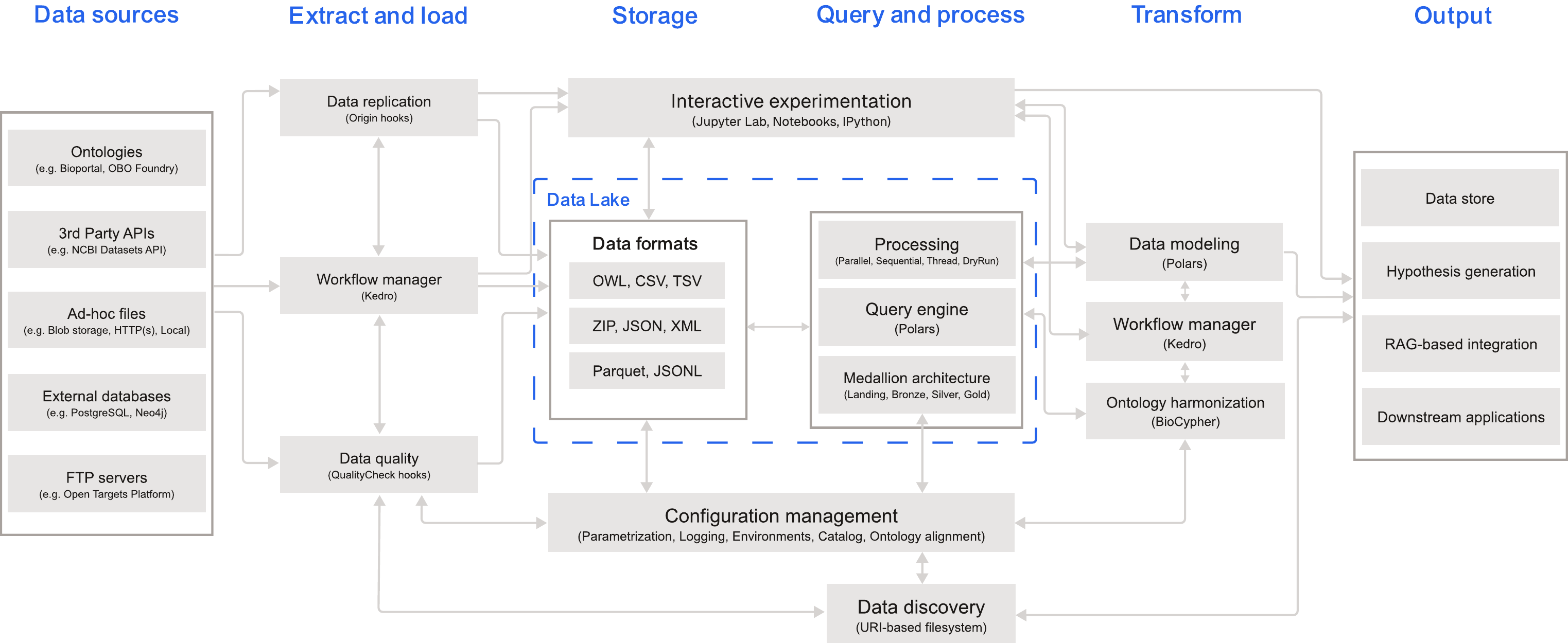}
  \caption{\textsbf{\kg data pipeline architecture.} Heterogeneous sources are ingested in the \textsc{Landing} layer via data replication and Kedro-managed workflows. Configuration management and URI-based data discovery primitives provide governance and traceability. The pipeline uses a medallion architecture to logically organize the data in increasing structure and quality as it flows through each layer (\textsc{Landing}, \textsc{Bronze}, \textsc{Silver}, and \textsc{Gold}). Transformations are done with Polars. The pipeline supports parallel, sequential, and thread execution modes. BioCypher~\cite{lobentanzer_democratizing_2023} is used to validate that each entity in the graph is ontology-grounded.
  }
  \label{fig:data-pipeline}
\end{figure}
\clearpage

\begin{figure}[!t]
  \centering
  \includegraphics[width=0.85\linewidth]{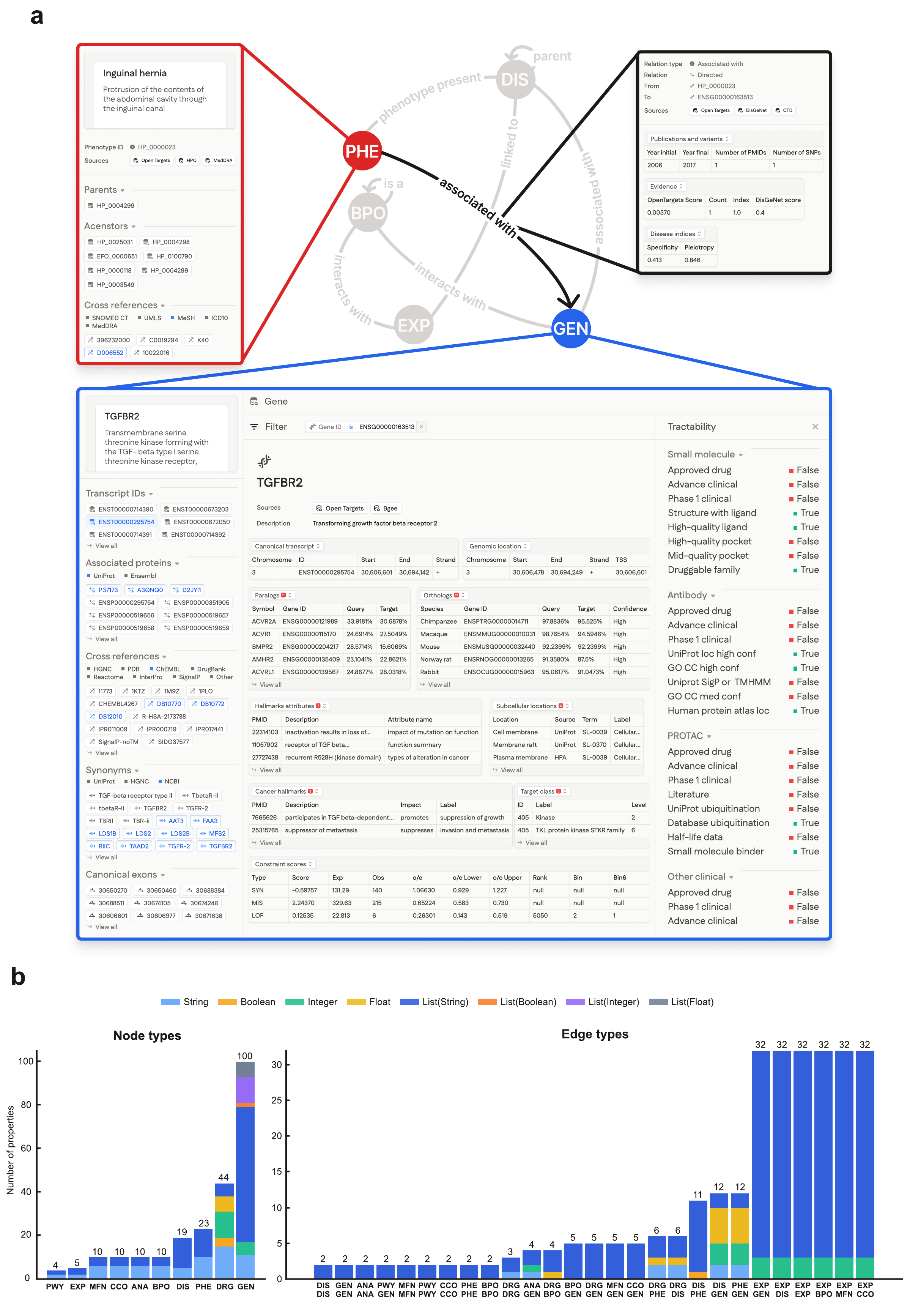}
  \caption{\revision{\textsbf{Property distribution in \kg.} 
  \textsbf{(a)} Representative subgraph from \kg illustrating a phenotype-gene association. A phenotype node (Inguinal hernia, HP\_0000023) is connected to a gene node (TGFBR2, ENSG00000163513) via an \textsc{ASSOCIATED-WITH} relation, with supporting evidence, provenance, and cross-references embedded as node and edge properties. \textsbf{(b)} Distribution of property types across node and edge types. Bar plots summarize the number and composition of properties for each entity across the graph.
  }}
  \label{fig:metadata}
\end{figure}
\clearpage

\begin{figure}[!t]
  \centering
  \includegraphics[width=0.8\linewidth]{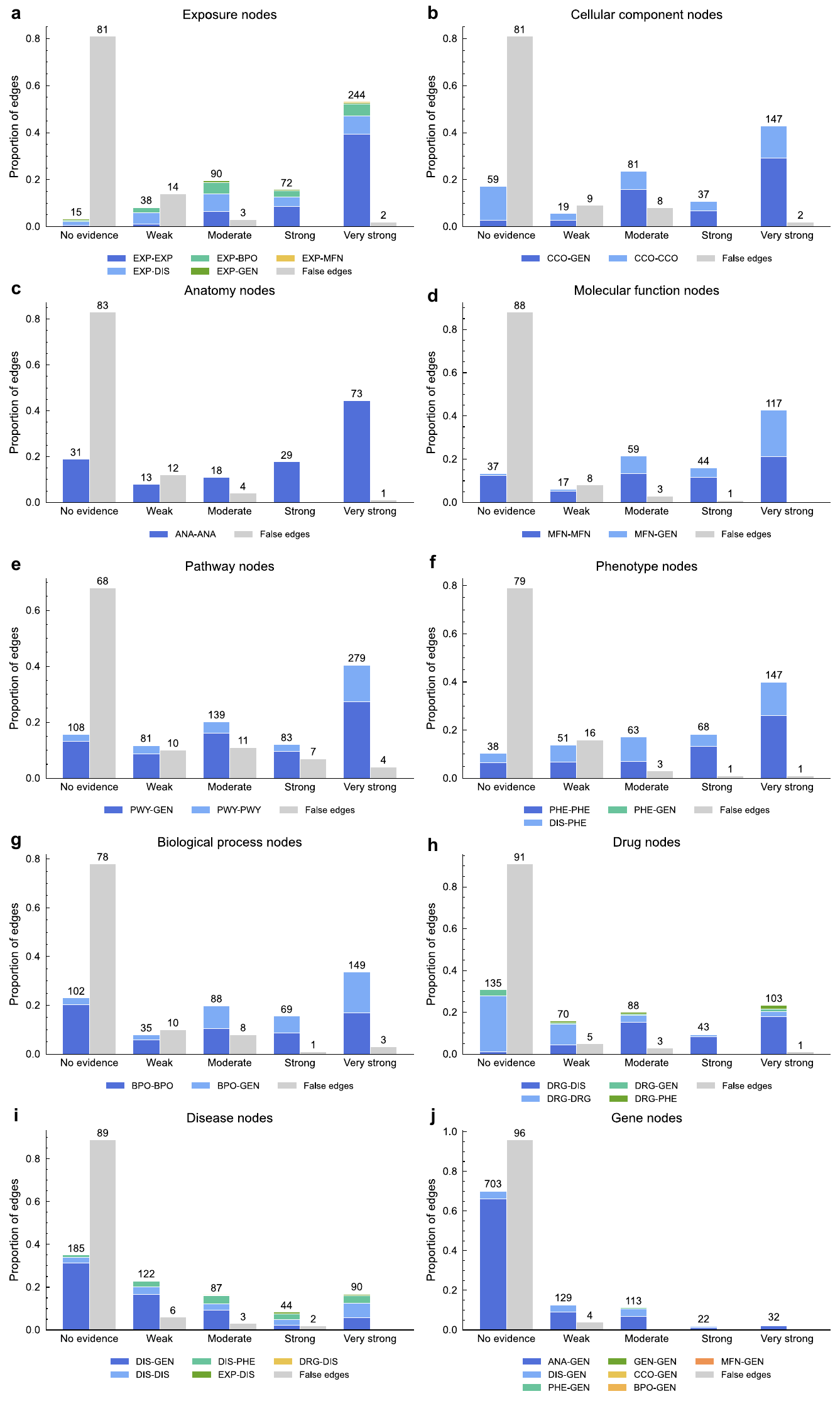}
  \caption{\textsbf{Validation of edges in \kg with a multimodal deep research agent.} Stacked bar plots}
  \label{fig:paperqa}
\end{figure}
\clearpage

\noindent
\begin{figure}[H]\ContinuedFloat
  \caption*{
  show the distribution of literature evidence scores (no evidence, weak, moderate, strong, very strong) for true edges (colored by edge type) and false edges (gray) for each node type.
  \textsbf{(a)} 459 true edges and 100 false edges incident on exposure (EXP) nodes were evaluated. Evidence was identified for 96.7\% of true edges ($n = 38$ weak, $90$ moderate, $72$ strong, $244$ very strong).
  \textsbf{(b)} 343 true edges and 100 false edges incident on cellular component (CCO) nodes were evaluated, with evidence identified for 82.8\% of true edges ($n = 19$ weak, $81$ moderate, $37$ strong, $147$ very strong).
  \textsbf{(c)} 164 true edges and 100 false edges incident on anatomy (ANA) nodes were evaluated, with evidence identified for 81.1\% of true edges ($n = 13$ weak, $18$ moderate, $29$ strong, $73$ very strong).
  \textsbf{(d)} 274 true edges and 100 false edges incident on molecular function (MFN) nodes were evaluated, with evidence identified for 86.5\% of true edges ($n = 17$ weak, $59$ moderate, $44$ strong, $117$ very strong).
  \textsbf{(e)} 690 true edges and 100 false edges incident on pathway (PWY) nodes were evaluated, with evidence identified for 84.3\% of true edges ($n = 81$ weak, $139$ moderate, $83$ strong, $279$ very strong).
  \textsbf{(f)} 367 true edges and 100 false edges incident on phenotype (PHE) nodes were evaluated, with evidence identified for 89.6\% of true edges ($n = 51$ weak, $63$ moderate, $68$ strong, $147$ very strong).
  \textsbf{(g)} 444 true edges and 100 false edges incident on biological process (BPO) nodes were evaluated, with evidence identified for 77.0\% of true edges ($n = 35$ weak, $88$ moderate, $69$ strong, $149$ very strong, $1$ failed evaluation). 
  \textsbf{(h)} 439 true edges and 100 false edges incident on drug (DRG) nodes were evaluated, with evidence identified for 69.2\% of true edges ($n = 70$ weak, $88$ moderate, $43$ strong, $103$ very strong).
  \textsbf{(i)} 528 true edges and 100 false edges incident on disease (DIS) nodes were evaluated, with evidence identified for 65.0\% of true edges ($n = 122$ weak, $87$ moderate, $44$ strong, $90$ very strong).
  \textsbf{(j)} 1,000 true edges and 100 false edges incident on gene (GEN) nodes were evaluated, with evidence identified for 29.6\% of true edges ($n = 129$ weak, $113$ moderate, $22$ strong, $32$ very strong, $1$ failed evaluation). Across all node types, false edges were predominantly assigned no evidence, supporting the specificity of the evaluation. 
  }
\end{figure}
\clearpage

\noindent
\begin{figure}[!t]
  \centering
  \includegraphics[width=\linewidth]{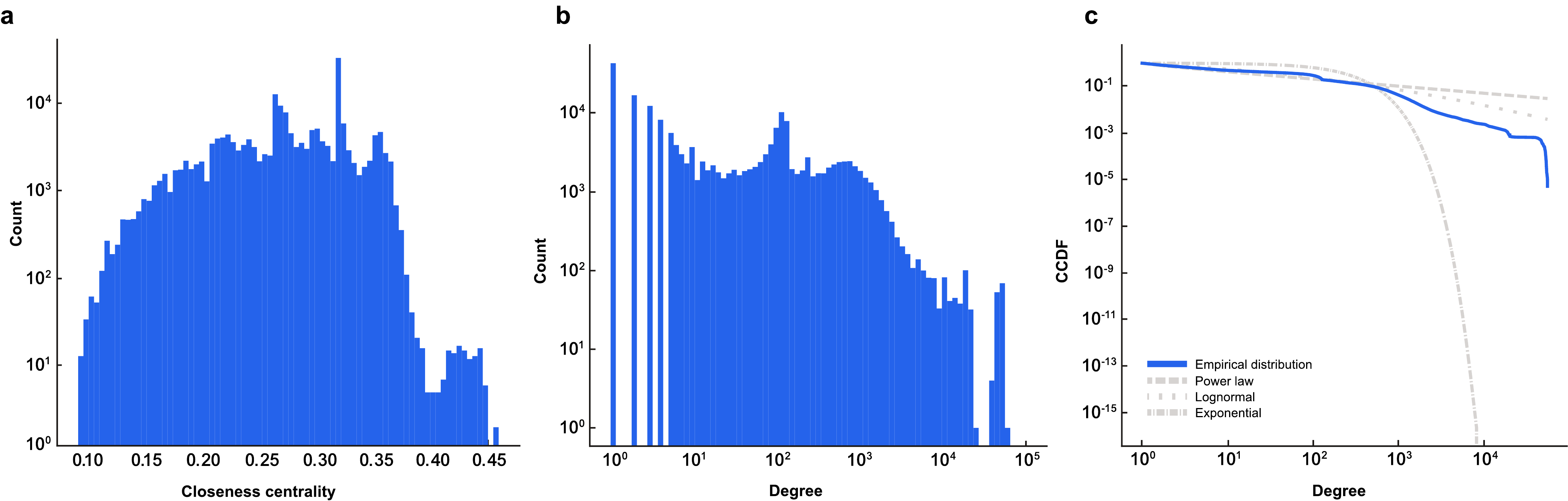}
  \caption{\revision{\textsbf{Topological properties of \kg.} 
  \textsbf{(a)} Closeness centrality is concentrated at intermediate values with a sparse tail of highly central nodes, indicating that most entities are separated by few hops, while a small subset serves as global hub nodes. \textsbf{(b)} Degree distribution (log-log) with a heavy tail and few high-degree hubs with many low-degree nodes, reflecting heterogeneous connectivity. \textsbf{(c)} Complementary cumulative distribution function (CCDF) of node degree with different fitted models shows slower-than-exponential decay, consistent with rich structure and long-range connectivity.}
  }
  \label{fig:topology}
\end{figure}

%% file: 300tables.tex
\begin{table}[h]
\centering
\small
\renewcommand{\arraystretch}{1.12}
\setlength{\tabcolsep}{6pt}
\rowcolors{2}{gray!8}{white}

\begin{tabular}{
>{\raggedright\arraybackslash}p{0.20\textwidth}
>{\centering\arraybackslash}p{0.06\textwidth}
>{\centering\arraybackslash}p{0.08\textwidth}
>{\centering\arraybackslash}p{0.12\textwidth}
>{\raggedright\arraybackslash}p{0.20\textwidth}
>{\raggedright\arraybackslash}p{0.2\textwidth}
}
\toprule
\rowcolor{gray!20}
\textbf{Node type} & \textbf{Label} & \textbf{Count} & \textbf{Percent (\%)} & \textbf{Primary sources} & \textbf{Ontologies} \\

Gene 
& GEN & 61,306 & 32.1
& Bgee, DisGeNET, Open Targets 
& ENSG, NCBIGene \\

Disease 
& DIS & 36,753 & 19.2 
& DrugCentral, Mondo, Open Targets 
& DOID, EFO, MONDO, MP, NCIT, OBA, OBI, OGMS, OTAR, Orphanet, PATO \\

Biological Process 
& BPO & 25,754 & 13.5
& GO 
& GO \\

Phenotype 
& PHE & 19,341 & 10.1 
& HPO, MedDRA, Open Targets 
& HP, MedDRA \\

Drug 
& DRG & 16,766 & 8.8 
& DrugBank, DrugCentral, OnSIDES, Open Targets 
& CHEMBL, DRUGBANK, RXNORM \\

Anatomy 
& ANA & 13,120 & 6.9
& Uberon 
& UBERON \\

Molecular Function 
& MFN & 10,161 & 5.3 
& GO 
& GO \\

Cellular Component 
& CCO & 4,052 & 2.1 
& GO 
& GO \\

Pathway 
& PWY & 2,805 & 1.5 
& Open Targets, Reactome 
& REACT \\

Exposure 
& EXP & 881 & 0.5 
& CTD 
& MESH \\

\midrule
\textbf{Total} 
& 
& \textbf{190,939} 
& \textbf{100.0} 
& \textbf{13} 
& \textbf{22} \\

\bottomrule
\end{tabular}

\caption{\textsbf{Distribution of node types in \kg.} Each node type in \kg is characterized by its provenance from primary data sources and, where applicable, by its alignment to standardized ontologies or controlled vocabularies. Primary sources denote resources ingested through the \kg pipeline, whereas ontology and vocabulary mappings provide harmonized identifiers and semantic standardization across nodes.}
\label{table:node-types}
\end{table}

\clearpage
\thispagestyle{empty}


\newlength{\edgecontentwidth}
\begingroup
\scriptsize
\renewcommand{\arraystretch}{1}
\setlength{\tabcolsep}{3pt}

\setlength{\edgecontentwidth}
  {\dimexpr\textwidth-10\tabcolsep\relax}

\setlength{\LTcapwidth}{\textwidth}
\setlength{\LTleft}{0pt}
\setlength{\LTright}{0pt}
\rowcolors{2}{gray!8}{white}

\begin{longtable}{@{}
>{\centering\arraybackslash}p{0.07\edgecontentwidth}
>{\centering\arraybackslash}p{0.08\edgecontentwidth}
>{\centering\arraybackslash}p{0.04\edgecontentwidth}
>{\raggedright\arraybackslash}p{0.40\edgecontentwidth}
>{\raggedright\arraybackslash}p{0.10\edgecontentwidth}
>{\raggedright\arraybackslash}p{0.30\edgecontentwidth}
@{}}

\toprule
\rowcolor{gray!20}
\textbf{Label} &
\textbf{Count} &
\textbf{(\%)} &
\textbf{Relation types} &
\textbf{Primary sources} &
\textbf{Indirect sources} \\
\midrule
\endfirsthead

\rowcolor{white}
\multicolumn{6}{@{}l}{
  \tablename~\thetable\ \textit{(continued)}
}\\
\toprule
\rowcolor{gray!20}
\textbf{Label} &
\textbf{Count} &
\textbf{(\%)} &
\textbf{Relation types} &
\textbf{Primary sources} &
\textbf{Indirect sources} \\
\midrule
\endhead

\midrule
\rowcolor{white}
\multicolumn{6}{r@{}}{
  \scriptsize\textit{Continued on next page}
}\\
\endfoot

\endlastfoot



DIS-GEN & 9,734,774 & 44.6 & \texttt{ASSOCIATED\_WITH} & DisGeNET, Open Targets & CGI, CTD, ClinGen, Genomics England, Orphanet, PsyGeNET, UniProt \\
ANA-GEN & 8,787,955 & 40.3 & \texttt{EXPRESSION\_PRESENT}, \texttt{EXPRESSION\_ABSENT} & Bgee & \\
DRG-DRG & 1,345,376 & 6.2 & \texttt{SYNERGISTIC\_INTERACTION}, \texttt{PARENT} & DrugBank, Open Targets & \\
PHE-GEN & 793,279 & 3.6 & \texttt{ASSOCIATED\_WITH} & DisGeNET, Open Targets & CTD, ClinGen, Genomics England, Orphanet, UniProt \\
GEN-GEN & 327,924 & 1.5 & \texttt{INTERACTS\_WITH} & PrimeKG & APID, BioGRID, BioPlex, CoFrac, ENCODE, HINT-binary, HINT-complex, HIPPIE, HiUnion, InSIDER, InSTRUCT, InWeb, InnateDB, IntAct, Interactome3D, KinomeNetX, LitBM17, MINT, PINA, PhosphoSP, QuBIC, SignaLink \\
BPO-GEN & 158,410 & 0.7 & \texttt{INTERACTS\_WITH} & Open Targets & \\
DIS-PHE & 157,144 & 0.7 & \texttt{PHENOTYPE\_PRESENT} & Open Targets & HPO \\
CCO-GEN & 105,309 & 0.5 & \texttt{INTERACTS\_WITH} & Open Targets & \\
MFN-GEN & 90,933 & 0.4 & \texttt{INTERACTS\_WITH} & Open Targets & \\
DRG-DIS & 70,380 & 0.3 & \texttt{INDICATION}, \texttt{CONTRAINDICATION}, \texttt{OFF\_LABEL\_USE} & DrugCentral, Open Targets & ATC, Clinical Trials, DailyMed, EMA, FDA, INN, USAN \\
DIS-DIS & 47,892 & 0.2 & \texttt{PARENT} & Open Targets & \\
PWY-GEN & 46,977 & 0.2 & \texttt{INTERACTS\_WITH} & Open Targets & Reactome \\
BPO-BPO & 44,494 & 0.2 & \texttt{IS\_A} & GO & \\
PHE-PHE & 26,121 & 0.1 & \texttt{PARENT} & HPO & \\
DRG-GEN & 20,694 & 0.1 & \texttt{ACTIVATOR}, \texttt{AGONIST}, \texttt{ALLOSTERIC\_ANTAGONIST}, \texttt{ANTAGONIST}, \texttt{BINDING\_AGENT}, \texttt{BLOCKER}, \texttt{CARRIER}, \texttt{DEGRADER}, \texttt{ENZYME}, \texttt{INHIBITOR}, \texttt{INVERSE\_AGONIST}, \texttt{MODULATOR}, \texttt{NEGATIVE\_ALLOSTERIC\_MODULATOR}, \texttt{NEGATIVE\_MODULATOR}, \texttt{OPENER}, \texttt{PARTIAL\_AGONIST}, \texttt{POSITIVE\_ALLOSTERIC\_MODULATOR}, \texttt{POSITIVE\_MODULATOR}, \texttt{RELEASING\_AGENT}, \texttt{STABILISER}, \texttt{SUBSTRATE}, \texttt{TARGET}, \texttt{TRANSPORTER} & DrugBank, Open Targets & BNF, Clinical Trials, DOI, DailyMed, EMA, Expert, FDA, HMA, ISBN, IUPHAR, KEGG, Other\tnote{*}, PMC, Patent, PubChem, PubMed, UniProt, Wikipedia \\
ANA-ANA & 17,082 & 0.1 & \texttt{PARENT} & Uberon & \\
DRG-PHE & 13,758 & 0.1 & \texttt{ADVERSE\_DRUG\_REACTION}, \texttt{ASSOCIATED\_WITH}, \texttt{CONTRAINDICATION}, \texttt{INDICATION}, \texttt{OFF\_LABEL\_USE} & DrugCentral, OnSIDES, Open Targets & ATC, Clinical Trials, DailyMed, EMA, FDA, INN, USAN \\
MFN-MFN & 12,587 & 0.1 & \texttt{IS\_A} & GO & \\
CCO-CCO & 4,639 & 0.0 & \texttt{IS\_A} & GO & \\
EXP-GEN & 2,989 & 0.0 & \texttt{INTERACTS\_WITH} & CTD, Open Targets & \\
PWY-PWY & 2,819 & 0.0 & \texttt{PARENT} & Reactome & \\
EXP-EXP & 2,443 & 0.0 & \texttt{PARENT} & CTD & \\
EXP-DIS & 2,391 & 0.0 & \texttt{LINKED\_TO} & CTD & Mondo \\
EXP-BPO & 2,260 & 0.0 & \texttt{INTERACTS\_WITH} & CTD & GO \\
DRG-BPO & 62 & 0.0 & \texttt{INDICATION} & DrugCentral, Open Targets & \\
EXP-MFN & 47 & 0.0 & \texttt{INTERACTS\_WITH} & CTD & GO \\
EXP-CCO & 13 & 0.0 & \texttt{INTERACTS\_WITH} & CTD & GO \\

\midrule
\textbf{Total} & \textbf{21,818,752} & \textbf{100.0} &
&  &  \\
\bottomrule
\hiderowcolors

\multicolumn{6}{@{}p{\textwidth}@{}}{%
  \footnotesize
  \revision{%
    \textsuperscript{*} ``Other'' is an upstream source label
    used by Open Targets. \kg preserves this label and normalizes
    it as an enumerated type.
  }%
}\\*
\caption{\textsbf{Distribution of edge types in \kg.}
Counts and proportions of each edge type in \kg, with corresponding
primary and indirect sources. Primary sources are resources ingested
through the \kg pipeline. Indirect sources are upstream resources
contributing information through primary sources, including when a
primary source aggregates or serves data originating elsewhere; these
are represented through provenance relationships.
\revision{The edge count distribution is imbalanced because upstream
resources differ in coverage. Percentages are rounded to one decimal
place.}}
\label{table:edge-types}\\

\end{longtable}

\endgroup




\begin{table}[h]
\centering
\footnotesize
\renewcommand{\arraystretch}{1.12}
\setlength{\tabcolsep}{6pt}
\rowcolors{2}{gray!8}{white}

\begin{threeparttable}
\begin{tabular}{
>{\raggedright\arraybackslash}p{0.18\linewidth}
>{\raggedright\arraybackslash}p{0.32\linewidth}
>{\centering\arraybackslash}p{0.14\linewidth}
>{\centering\arraybackslash}p{0.1\linewidth}
>{\centering\arraybackslash}p{0.15\linewidth}
}
\toprule
\rowcolor{gray!20}
\textbf{Primary source} & \textbf{Indirect sources} & \textbf{Version} & \textbf{Format} & \textbf{License} \\

Open Targets 
& ATC, BNF, CGI, ClinGen, Clinical Trials, DailyMed, DOI, EMA, Expert, FDA, Genomics England, HMA, HPO, INN, ISBN, IUPHAR, KEGG, Orphanet, Other\tnote{*}, Patent, PMC, PsyGeNET, PubChem, PubMed, Reactome, UniProt, USAN, Wikipedia 
& 25.06 & Parquet & CC0 1.0 \\

DrugBank 
& BNF, Clinical Trials, DailyMed, DOI, EMA, Expert, FDA, HMA, ISBN, IUPHAR, KEGG, Other\revision{\tnote{*}}, Patent, PMC, PubChem, PubMed, UniProt, Wikipedia 
& 5.1.13 & XML, CSV & CC BY-NC 4.0 \\

DrugCentral 
& ATC, Clinical Trials, DailyMed, EMA, FDA, USAN 
& v2023-11-01 & SQL dump & CC BY-SA 4.0 \\

DisGeNET 
& CGI, ClinGen, Genomics England, Orphanet, PsyGeNET, UniProt 
& --- & TSV & Custom DisGeNET License. Similar to CC BY-NC 4.0. Commercial license available separately. \\

Bgee 
& 
& current & ZIP & CC0 1.0 \\

CTD 
& GO, MeSH, Mondo 
& current & ZIP & Noncommercial research and educational use \\

OnSIDES 
& MedDRA, RxNorm 
& v3.1.0 & ZIP & MIT \\

Reactome 
& 
& current & TSV & CC0 1.0 \\

HGNC 
& 
& current & TSV & CC0 1.0 \\

PrimeKG 
& APID, BioGRID, BioPlex, CoFrac, ENCODE, HiUnion, HINT-binary, HINT-complex, HIPPIE, InnateDB, InSIDER, InSTRUCT, IntAct, Interactome3D, InWeb, KinomeNetX, LitBM17, MINT, PhosphoSP, PINA, QuBIC, SignaLink 
& current & CSV & CC0 1.0 \\

\bottomrule
\end{tabular}

\begin{tablenotes}[flushleft]
\footnotesize
\revision{\item[*] ``Other'' is an upstream source label used by Open Targets. \kg preserves this label and normalizes it as an enumerated type.}
\end{tablenotes}

\caption{\textsbf{Data sources used in \kg construction.} \kg integrates primary sources that are ingested through the \kg pipeline. Beyond these data resources, \kg also incorporates indirect sources that are referenced through provenance relationships within the primary sources. This approach captures the full lineage of the underlying data for transparent tracing of each \kg association back to its original datasets, databases, and publications.}
\label{table:data-sources}
\end{threeparttable}
\end{table}

\clearpage
\thispagestyle{empty}

\begingroup

\small
\renewcommand{\arraystretch}{1.12}
\setlength{\LTpost}{0pt} 
\setlength{\tabcolsep}{6pt}
\rowcolors{2}{gray!8}{white}

\begin{xltabular}{\textwidth}{
    >{\raggedright\arraybackslash}p{0.10\textwidth}
    >{\raggedright\arraybackslash}p{0.50\textwidth}
    >{\raggedright\arraybackslash}X
}
\toprule
\rowcolor{gray!20}
\textbf{Source} &
\textbf{Landing dataset} &
\textbf{Access page} \\
\midrule
\endfirsthead

\toprule
\rowcolor{gray!20}
\textbf{Source} &
\textbf{Landing dataset} &
\textbf{Access page} \\
\midrule
\endhead

\textbf{Bgee} &
\path{bgee.homo_sapiens_expressions_advanced} &
\href{https://www.bgee.org/ftp/current/download/calls/expr_calls/Homo_sapiens_expr_advanced.tsv.gz}
{Homo sapiens expression calls} \\

\textbf{CTD} &
\path{ctd.ctd_exposure_events} &
\href{https://ctdbase.org/reports/CTD_exposure_events.csv.gz}
{CTD exposure events} \\

\textbf{DisGeNET} &
\path{disgenet.curated_gene_disease_associations} &
\href{https://web.archive.org/web/20240503020023/https://www.disgenet.org/static/disgenet_ap1/files/downloads/readme.txt}
{DisGeNET curated disease associations} \\

\textbf{DrugBank} &
\path{drugbank.full_database} &
\href{https://go.drugbank.com/releases/5-1-13/downloads/full-database}
{DrugBank database release 5-1-13} \\

\textbf{HGNC} &
\path{drugbank.gene_map} &
\href{https://www.genenames.org/cgi-bin/download/custom?col=md_eg_id\&col=md_prot_id\&status=Approved\&hgnc_dbtag=on\&order_by=gd_app_sym_sort\&format=text\&submit=submit}
{HGNC custom DrugBank gene map download} \\

\textbf{DrugBank} &
\path{drugbank.carrier} &
\href{https://go.drugbank.com/releases/5-1-13/downloads/carrier}
{DrugBank carrier release 5-1-13} \\

\textbf{DrugBank} &
\path{drugbank.enzyme} &
\href{https://go.drugbank.com/releases/5-1-13/downloads/enzyme}
{DrugBank enzyme release 5-1-13} \\

\textbf{DrugBank} &
\path{drugbank.target} &
\href{https://go.drugbank.com/releases/5-1-13/downloads/target}
{DrugBank target release 5-1-13} \\

\textbf{DrugBank} &
\path{drugbank.transporter} &
\href{https://go.drugbank.com/releases/5-1-13/downloads/transporter}
{DrugBank transporter release 5-1-13} \\

\textbf{DrugBank} &
\path{drugbank.vocabulary} &
\href{https://go.drugbank.com/releases/5-1-13/downloads/all-drugbank-vocabulary}
{DrugBank vocabulary release 5-1-13} \\

\textbf{DrugCentral} &
\path{drugcentral.psql_dump} &
\href{https://unmtid-dbs.net/download/drugcentral.dump.11012023.sql.gz}
{DrugCentral PostgreSQL dump} \\

\textbf{HGNC} &
\path{gene_names.gene_names} &
\href{https://www.genenames.org/cgi-bin/download/custom?col=gd_app_sym\&col=gd_app_name\&col=gd_pub_acc_ids\&col=gd_pub_refseq_ids\&col=gd_pub_eg_id\&col=md_eg_id\&col=md_prot_id\&col=md_mim_id\&status=Approved\&hgnc_dbtag=on\&order_by=gd_app_sym_sort\&format=text\&submit=submit}
{HGNC custom gene names download} \\



\textbf{OnSIDES} &
\path{onsides.high_confidence} &
\href{https://github.com/tatonetti-lab/onsides/releases/download/v3.1.0/onsides-v3.1.0.zip}
{OnSIDES v3.1.0 archive} \\

\textbf{OnSIDES} &
\path{onsides.vocab_meddra_adverse_effect} &
\href{https://github.com/tatonetti-lab/onsides/releases/download/v3.1.0/onsides-v3.1.0.zip}
{OnSIDES v3.1.0 archive} \\

\textbf{OnSIDES} &
\path{onsides.vocab_rxnorm_ingredient} &
\href{https://github.com/tatonetti-lab/onsides/releases/download/v3.1.0/onsides-v3.1.0.zip}
{OnSIDES v3.1.0 archive} \\

\textbf{Open Targets} &
\path{opentargets.disease} &
\href{ftp://ftp.ebi.ac.uk/pub/databases/opentargets/platform/25.06/output/disease}
{Open Targets 25.06 disease} \\

\textbf{Open Targets} &
\path{opentargets.disease_phenotype} &
\href{ftp://ftp.ebi.ac.uk/pub/databases/opentargets/platform/25.06/output/disease_phenotype}
{Open Targets 25.06 disease phenotype} \\

\textbf{Open Targets} &
\path{opentargets.drug_indication} &
\href{ftp://ftp.ebi.ac.uk/pub/databases/opentargets/platform/25.06/output/drug_indication}
{Open Targets 25.06 drug indication} \\

\textbf{Open Targets} &
\path{opentargets.drug_mechanism_of_action} &
\href{ftp://ftp.ebi.ac.uk/pub/databases/opentargets/platform/25.06/output/drug_mechanism_of_action}
{Open Targets 25.06 drug mechanism of action} \\

\textbf{Open Targets} &
\path{opentargets.drug_molecule} &
\href{ftp://ftp.ebi.ac.uk/pub/databases/opentargets/platform/25.06/output/drug_molecule}
{Open Targets 25.06 drug molecule} \\

\textbf{Open Targets} &
\path{opentargets.target} &
\href{ftp://ftp.ebi.ac.uk/pub/databases/opentargets/platform/25.06/output/target}
{Open Targets 25.06 target} \\

\textbf{Open Targets} &
\path{opentargets.target_disease_associations} &
\href{ftp://ftp.ebi.ac.uk/pub/databases/opentargets/platform/25.06/output/association_by_overall_indirect}
{Open Targets 25.06 association by overall indirect} \\

\textbf{PrimeKG} &
\path{ppi.protein_protein} &
\href{https://www.science.org/doi/10.1126/science.1257601}
{Menche, J. et al.} \\

\textbf{Reactome} &
\path{reactome.reactome_pathways} &
\href{https://reactome.org/download/current/ReactomePathways.txt}
{Reactome pathways} \\

\textbf{Reactome} &
\path{reactome.reactome_pathways_relation} &
\href{https://reactome.org/download/current/ReactomePathwaysRelation.txt}
{Reactome pathway relations} \\

\bottomrule

\caption{
\textsbf{Access links for the primary datasets used in \kg.}
The table lists the primary data sources incorporated into the \kg pipeline,
their corresponding landing dataset identifiers, and the locations from which
the source data or associated documentation can be accessed.%
}
\label{table:primary-source-access-links}
\end{xltabular}
\endgroup

\begin{table}[h]
\centering
\small
\renewcommand{\arraystretch}{1}
\setlength{\tabcolsep}{6pt}
\rowcolors{2}{gray!8}{white}

\begin{tabular}{
>{\raggedright\arraybackslash}p{0.2\textwidth}
>{\centering\arraybackslash}p{0.10\textwidth}
>{\raggedright\arraybackslash}p{0.30\textwidth}
>{\centering\arraybackslash}p{0.20\textwidth}
>{\centering\arraybackslash}p{0.08\textwidth}
}
\toprule
\rowcolor{gray!20}
\textbf{Source} & \textbf{Namespace} & \textbf{Description} & \textbf{Version} & \textbf{Format} \\

\rowcolor{gray!15}
\multicolumn{5}{l}{\textit{Explicitly ingested}} \\

Biolink Model 
& --- 
& Upper-level schema for biomedical entities and relations 
& v3.2.1  & OWL \\

Gene Ontology
& GO 
& Ontology of biological processes, cellular components, and molecular functions 
& current & JSON \\

Human Phenotype Ontology 
& HP 
& Vocabulary for human phenotypic abnormalities  
& v2025-05-06 & JSON \\

Uberon 
& UBERON 
& Cross-species anatomy ontology 
& v2026-04-01 & JSON \\

Mondo 
& MONDO 
& Unified disease ontology across classification systems 
& current & JSON \\

Disease Ontology 
& DOID 
& Structured disease classification for cross-resource mapping 
& v656 & OWL \\

Orphanet
& Orphanet 
& Rare disease ontology with clinical annotations 
& v30 & OWL \\

\rowcolor{gray!15}
\multicolumn{5}{l}{\textit{Referenced via Open Targets v25.06}} \\

Experimental Factor Ontology
& EFO
& Ontology of experimental variables and disease traits
& --- & OWL \\

Ontology of Biological Attributes
& OBA
& Ontology linking phenotypic attributes to entities
& --- & OWL \\

Open Targets Disease Ontology
& OTAR
& Internal disease classification used in Open Targets
& --- & JSON \\

NCI Thesaurus
& NCIT
& Cancer and biomedical terminology ontology
& --- & OWL \\

Phenotype and Trait Ontology
& PATO
& Ontology of phenotypic qualities
& --- & OWL \\

Mammalian Phenotype Ontology
& MP
& Vocabulary for mammalian phenotype annotation
& --- & OWL \\

Ontology for General Medical Science
& OGMS
& Ontology for clinical and disease representation
& --- & OWL \\

Ontology for Biomedical Investigations
& OBI
& Ontology of experimental protocols and assays
& --- & OWL \\

\rowcolor{gray!15}
\multicolumn{5}{l}{\textit{Controlled vocabularies}} \\

Medical Subject Headings
& MESH
& Thesaurus for indexing biomedical literature
& --- & UMLS \\

MedDRA
& MEDDRA
& Regulatory terminology for diseases and adverse events
& v3.1.0 & UMLS \\

RxNorm
& RXNORM
& Normalized drug ingredient vocabulary
& v3.1.0 & UMLS \\

\bottomrule
\end{tabular}

\caption{\textsbf{Ontologies and controlled vocabularies used to align entities in \kg.} Ontologies and controlled vocabularies are used to align \kg entities across data sources by providing standardized identifiers and semantic definitions. They are stratified by ingestion level, including ontologies ingested into \kg, ontologies referenced through upstream data sources, and external controlled vocabularies used for normalization and harmonization.}
\label{table:ontologies}
\end{table}

\begin{table}[htbp]
\centering
\small
\renewcommand{\arraystretch}{1.12}
\setlength{\tabcolsep}{6pt}
\rowcolors{2}{gray!8}{white}

\begin{tabularx}{\textwidth}{
    >{\raggedright\arraybackslash}p{0.24\linewidth}
    >{\raggedright\arraybackslash}p{0.30\linewidth}
    >{\raggedright\arraybackslash}X
}
\toprule
\rowcolor{gray!20}
\textbf{Ontology} &
\textbf{Landing dataset} &
\textbf{Download link} \\

\textbf{Biolink Model} &
\path{ontology.biolink} &
\href{https://data.bioontology.org/ontologies/BIOLINK/submissions/68/download?apikey=168d52c1-bd36-4866-aa99-0aa8eb06f295}
{BioPortal BIOLINK submission 68} \\

\textbf{Disease Ontology} &
\path{ontology.disease} &
\href{https://data.bioontology.org/ontologies/DOID/submissions/656/download?apikey=168d52c1-bd36-4866-aa99-0aa8eb06f295}
{BioPortal DOID submission 656} \\

\textbf{Gene Ontology} &
\path{ontology.go_plus} &
\href{https://purl.obolibrary.org/obo/go/extensions/go-plus.json}
{GO Plus JSON} \\

\textbf{Human Phenotype Ontology} &
\path{ontology.hp} &
\href{https://github.com/obophenotype/human-phenotype-ontology/releases/download/v2025-05-06/hp-full.json}
{HPO full JSON v2025-05-06} \\

\textbf{Mondo Disease Ontology} &
\path{ontology.mondo} &
\href{http://purl.obolibrary.org/obo/mondo.json}
{Mondo JSON} \\

\textbf{Orphanet ORDO} &
\path{ontology.orphanet} &
\href{https://data.bioontology.org/ontologies/ORDO/submissions/30/download?apikey=168d52c1-bd36-4866-aa99-0aa8eb06f295}
{BioPortal ORDO submission 30} \\

\textbf{Uberon} &
\path{ontology.uberon} &
\href{https://github.com/obophenotype/uberon/releases/download/v2026-04-01/human-view.json}
{Uberon human-view JSON v2026-04-01} \\

\bottomrule
\end{tabularx}

\caption{\textbf{Download links for the ontologies used in \kg.}
The table lists the ontology datasets incorporated into the \kg pipeline,
together with their corresponding landing dataset identifiers and download
locations.}
\label{table:ontology-download-links}
\end{table}

\begin{table}[h]
\centering
\footnotesize
\renewcommand{\arraystretch}{1.15}
\setlength{\tabcolsep}{5pt}
\rowcolors{2}{gray!6}{white}

\begin{tabular}{
>{\raggedright\arraybackslash}p{0.22\linewidth}
>{\raggedright\arraybackslash}p{0.40\linewidth}
>{\raggedright\arraybackslash}p{0.28\linewidth}
}
\toprule
\rowcolor{gray!15}
\textbf{Dataset name} & \textbf{Description} & \textbf{Source} \\
\midrule

ParquetDataset 
& Loads and saves Parquet files using Polars. Supports local, remote and embedded filesystems via \code{fsspec}. Includes schema-based type parsing via configuration. 
& \code{optimuskg.datasets} \\

JsonDataset 
& Loads and saves (ND)JSON files using Polars. 
& \code{optimuskg.datasets} \\

ZipDataset 
& Wraps a zip archive and extracts a file to load via a configurable nested dataset type. 
& \code{optimuskg.datasets} \\

SQLDumpQueryDataset 
& Restores a PostgreSQL database from a gzipped SQL dump using Docker, executes a SQL query, and returns a Polars DataFrame.
& \code{optimuskg.datasets} \\

LXMLDataset 
& Loads and saves XML files using \code{lxml.etree}.
& \code{optimuskg.datasets} \\

OWLDataset 
& Loads and saves OWL files (XML-based). Returns a dataclass containing the content string and filepath. 
& \code{optimuskg.datasets} \\

CSVDataset 
& Loads and saves CSV files using Polars.
& \code{optimuskg.datasets} \\

PartitionedDataset 
& Manages collections of partitioned files as a single logical dataset.
& \code{kedro\_datasets} \\

\bottomrule
\end{tabular}

\caption{\textsbf{Dataset abstractions in the \kg pipeline.} Each dataset abstraction provides a standardized interface for loading and persisting data across storage backends. Most catalog entries (77.9\%) use the custom \texttt{ParquetDataset}, whereas the remaining entries rely on Kedro built-in abstractions, primarily \texttt{PartitionedDataset}.}
\label{table:dataset-abstractions}
\end{table}

\begin{table}[t]
\centering
\small
\renewcommand{\arraystretch}{1.15}
\setlength{\tabcolsep}{6pt}
\rowcolors{2}{gray!6}{white}

\begin{tabular}{
>{\raggedright\arraybackslash}p{0.18\linewidth}
>{\raggedright\arraybackslash}p{0.24\linewidth}
>{\raggedright\arraybackslash}p{0.46\linewidth}
}
\toprule
\rowcolor{gray!15}
\textbf{Hook} & \textbf{Layers} & \textbf{Behavior} \\
\midrule

\texttt{Origin} & \textsc{Landing} &
Reads the \texttt{origin} metadata field from the catalog entry and validates it as a known provider model (HTTP, DrugBank, BioOntology, Open Targets). If the file does not exist, triggers a download via the appropriate provider. If origin metadata is missing and the file is absent, creates an empty placeholder file with the correct schema to prevent pipeline failure on private or manual datasets. \\

\midrule

\texttt{Checksum} & \textsc{Landing, Bronze, Silver} &
Before loading any dataset, computes a Blake2b checksum of the file on disk and compares it against the expected value stored in \texttt{metadata.checksum}. Logs a warning on mismatch. \\

\midrule

\texttt{QualityChecks} & \textsc{Silver, Gold} &
Performs post-node validation and logs warnings for violations. Enforces \texttt{snake\_case} naming conventions for all columns. Validates that columns prefixed with \texttt{id} contain no null values. Ensures that values in the \texttt{relation} column conform to the \texttt{Relation} enum defined in \texttt{constants.py}, raising a \texttt{DatasetError} for unknown values and logging warnings when mapped to \texttt{Relation.OTHER}, including counts and proportions of affected rows. \\

\bottomrule
\end{tabular}

\caption{
\textsbf{Overview of Kedro hooks and dataset-level validation mechanisms across pipeline layers in \kg.} The table summarizes hooks that operate at different stages of the \kg pipeline and their corresponding behaviors. Hooks are executed in a last-in, first-out order: \texttt{Origin} and \texttt{Checksum} are applied prior to pipeline execution during dataset loading, whereas \texttt{QualityChecks} are applied after node execution. 
The \texttt{Origin} hook ensures that input data are available by validating the declared data provider and triggering downloads when needed, or by creating schema-compliant placeholder files for missing private or manually curated datasets. The \texttt{Checksum} hook verifies data integrity by comparing file checksums against expected values stored in metadata. The \texttt{QualityChecks} hook performs post-processing validation, enforcing naming conventions, checking identifier completeness, and validating relation types against controlled vocabularies.
Hard failures (for example, invalid relation values or missing required files) halt pipeline execution, whereas soft warnings (for example, checksum mismatches, naming violations, or null identifiers) are logged but do not interrupt execution. In addition, schema enforcement is applied at dataset load time within the dataset abstraction, ensuring type consistency at ingestion.
}
\label{tab:kedro_hooks}
\end{table}

\begin{table}[h]
\centering
\small
\renewcommand{\arraystretch}{1.12}
\setlength{\tabcolsep}{6pt}
\rowcolors{2}{gray!8}{white}

\begin{tabular}{
>{\raggedright\arraybackslash}p{0.18\linewidth}
>{\centering\arraybackslash}p{0.14\linewidth}
>{\raggedright\arraybackslash}p{0.56\linewidth}
}
\toprule
\rowcolor{gray!20}
\textbf{Key} & \textbf{Type} & \textbf{Description} \\

id 
& String 
& Globally unique node identifier in CURIE format \texttt{\{namespace\}[:\_]\{local\_id\}}. Namespace prefixes map to source ontologies (e.g., ENSG, GO, HP, CHEMBL) and may share node types (e.g., EFO and MONDO map to DIS). All prefixes are validated against the Biolink Model (\textit{e.g.}, ENSG00000139618, GO\_0000001, MESH:C000188).
 \\

label 
& String 
& Node type label matching the enumerated 3-letter values  (\eg, GEN, DIS, DRG) \\

properties 
& Struct 
& Nested struct containing all node-type-specific attributes (varies per node type) \\

\bottomrule
\end{tabular}

\caption{\textsbf{Top-level schema shared across all node types in \kg.} Each node is represented by a unique identifier, a type label, and a \texttt{properties} struct that contains attributes specific to the node type.}
\label{table:node-schema}
\end{table}

\begin{table}[h]
\centering
\small
\renewcommand{\arraystretch}{1.12}
\setlength{\tabcolsep}{6pt}
\rowcolors{2}{gray!8}{white}

\begin{tabular}{
>{\raggedright\arraybackslash}p{0.18\linewidth}
>{\centering\arraybackslash}p{0.14\linewidth}
>{\raggedright\arraybackslash}p{0.56\linewidth}
}
\toprule
\rowcolor{gray!20}
\textbf{Key} & \textbf{Type} & \textbf{Description} \\

from 
& String 
& Source node ID \\

to 
& String 
& Target node ID \\

label 
& String 
& Edge type label encoding the node type pair in \texttt{\{from\}-\{to\}} format (\eg, ANA-GEN, DRG-DIS) \\

relation 
& String 
& Semantic relation type with enumerated values (\eg, EXPRESSION\_PRESENT, INDICATION) \\

undirected 
& Boolean 
& Whether the edge is undirected or directed\\

properties 
& Struct 
& Nested struct containing edge-type-specific attributes (varies per edge type) \\

\bottomrule
\end{tabular}

\caption{\textsbf{Top-level schema shared across all edge types in \kg.} Each edge is defined by a source node and a target node, together with a typed relation that specifies the nature of the connection. Additional edge-specific attributes are stored in a \texttt{properties} struct.}
\label{table:edge-schema}
\end{table}

\begin{table}[h]
\centering
\small
\renewcommand{\arraystretch}{1.12}
\setlength{\tabcolsep}{6pt}
\rowcolors{2}{gray!8}{white}

\begin{tabular}{
>{\raggedright\arraybackslash}p{0.25\linewidth}
>{\centering\arraybackslash}p{0.1\linewidth}
}
\toprule
\rowcolor{gray!20}
\textbf{Node type} & \textbf{Label} \\

Gene               & GEN \\
Anatomy            & ANA \\
Disease            & DIS \\
Drug               & DRG \\
Environmental exposure           & EXP \\
Phenotype          & PHE \\
Pathway            & PWY \\
Biological process & BPO \\
Cellular component & CCO \\
Molecular function & MFN \\

\bottomrule
\end{tabular}

\caption{\textsbf{Mapping of node types to node type labels in \kg.} Each node type is assigned a standardized 3-letter label used throughout \kg. Edge labels are generated dynamically from pairs of node labels using the format \texttt{\{from\}-\{to\}} (\eg, ANA-GEN, DRG-DIS).}
\label{table:node-labels}
\end{table}

\begin{table}[h]
\centering
\small
\renewcommand{\arraystretch}{1.12}
\setlength{\tabcolsep}{6pt}
\rowcolors{2}{gray!8}{white}

\begin{tabular}{
>{\raggedright\arraybackslash}p{0.2\linewidth}
>{\centering\arraybackslash}p{0.2\linewidth}
}
\toprule
\rowcolor{gray!20}
\textbf{Size (nodes)} & \textbf{Number of components} \\

189,774 & 1 \\
226     & 1 \\
6       & 1 \\
5       & 6 \\
4       & 3 \\
3       & 55 \\
2       & 363 \\

\bottomrule
\end{tabular}
\caption{\revision{\textsbf{Distribution of connected network components in \kg.} The largest connected component (LCC) contains 189,774 nodes (approximately 99.39\% of the graph), whereas the remaining 1,165 nodes are distributed across 429 smaller components, primarily pairs and triplets.}}
\label{table:connected-components}
\end{table}